\documentclass[lettersize,journal]{IEEEtran}
\usepackage{cite}
\usepackage{amsmath,amssymb,amsfonts}
\usepackage{graphicx}
\usepackage[bottom]{footmisc}
\usepackage[parskip=20pt,captionskip=0pt]{subfig}
\usepackage{textcomp}
\usepackage[dvipsnames]{xcolor}
\def\BibTeX{{\rm B\kern-.05em{\sc i\kern-.025em b}\kern-.08em
    T\kern-.1667em\lower.7ex\hbox{E}\kern-.125emX}}
\usepackage{url}
\usepackage{fancyhdr,amsmath,graphicx} 
\usepackage{fontenc}    
\usepackage{xcolor}         
\usepackage{url}            
\usepackage{amsfonts}       
\usepackage{nicefrac}       
\usepackage{microtype}      

\usepackage[square,sort,comma,numbers]{natbib}
\usepackage{colortbl}
\usepackage{multirow,makecell}
\usepackage{algpseudocode}
\usepackage{graphics}
\usepackage{epsfig}
\usepackage{xfrac}
\usepackage[ruled,vlined,linesnumbered]{algorithm2e}
\usepackage{color}
\usepackage{float}
\usepackage{multicol}
\usepackage{circledsteps}
\pgfkeys{/csteps/inner color=white}
\pgfkeys{/csteps/outer color=white}
\pgfkeys{/csteps/fill color=black}
\pgfkeys{/csteps/inner ysep=3pt}
\pgfkeys{/csteps/inner xsep=3pt}
\usepackage{xspace}
\usepackage{wrapfig,lipsum,booktabs}
\usepackage{enumitem}

\input{mysymbol.sty}
\definecolor{mygray}{gray}{0.9}
\newcolumntype{a}{>{\columncolor{mygray}}c}

\usepackage[colorlinks = true,
            linkcolor = Violet,
            urlcolor  = Violet,
            citecolor = Violet]{hyperref}       

\newcommand{\method}{{\footnotesize{\textsf{GCond}}\xspace}}
\newcommand{{\methodpca}}{{\footnotesize{\textsf{GCond-PCA}\xspace}}}
\newcommand{{\methodica}}{{\footnotesize{\textsf{GCond-ICA}}\xspace}}
\newcommand{{\methodlda}}{{\footnotesize{\textsf{GCond-LDA}}\xspace}}
\newcommand{{\methodvgae}}{{\footnotesize{\textsf{GCond-VGAE}}\xspace}}

\newcommand{\graphpca}{{\footnotesize{\textsf{GraphPCA}}\xspace}}
\newcommand{\tinygraphlin}{{\footnotesize{\textsf{TinyGraph-Lin}\xspace}}}

\newcommand{\tinygraphthree}{{\footnotesize{\textsf{TinyGraph-l3}\xspace}}}
\newcommand{\tinygraphhfot}{{\footnotesize{\textsf{TinyGraph-h512}\xspace}}}
\newcommand{\tinygraphote}{{\footnotesize{\textsf{TinyGraph-h128}\xspace}}}
\newcommand{\methodpcaf}{{\footnotesize{\textsf{GCond-PCA-f}\xspace}}}
\newcommand{{\methodicaf}}{{\footnotesize{\textsf{GCond-ICA-f}}\xspace}}
\newcommand{{\methodldaf}}{{\footnotesize{\textsf{GCond-LDA-f}}\xspace}}
\newcommand{\tinygraphmlp}{{\footnotesize{\textsf{TinyGraph-MLP}\xspace}}}
\newcommand{\tinygraphgcn}{{\footnotesize{\textsf{TinyGraph-GCN}\xspace}}}
\newcommand{\tinygraphx}{{\footnotesize{\textsf{TinyGraph-X}\xspace}}}
\newcommand{\tinygraph}{{\footnotesize{\textsf{TinyGraph}}}\xspace}
\newcommand{\tinygraphone}{{\footnotesize{\textsf{TinyGraph-One}}}\xspace}

\definecolor{myred}{HTML}{E33226}

\newcommand{\cora}{{\small\texttt{Cora}}\xspace}
\newcommand{\citeseer}{{\small\texttt{Citeseer}}\xspace}
\newcommand{\arxiv}{{\small\texttt{Arxiv}}\xspace}
\newcommand{\flickr}{{\small\texttt{Flickr}}\xspace}
\newcommand{\reddit}{{\small\texttt{Reddit}}\xspace}

\usepackage[capitalize,noabbrev,nameinlink]{cleveref}
\usepackage[colorinlistoftodos]{todonotes}

\title{{\textsf{TinyGraph}: Joint Feature and Node Condensation for Graph Neural Networks}}
\author{Yezi Liu \quad Yanning Shen
\IEEEcompsocitemizethanks{Yezi Liu and Yanning Shen are with the Department of Electrical Engineering and Computer Science, University of California Irvine, Irvine, CA 92623 USA (e-mails: yezil3@uci.edu, yannings@uci.edu). 
Corresponding author (yannings@uci.edu).}
}
\begin{document}

\maketitle

\begin{abstract} 
Training graph neural networks (GNNs) on large-scale graphs can be challenging due to the high computational expense caused by the massive number of nodes and high-dimensional nodal features. 
Existing graph condensation studies tackle this problem only by reducing the number of nodes in the graph. However, the resulting condensed graph data can still be cumbersome. Specifically, although the nodes of the \citeseer dataset are reduced to $0.9\%$ ($30$ nodes) in training, the number of features is $3,703$, severely exceeding the training sample magnitude. Faced with this challenge, we study the problem of \textit{joint condensation} for both features and nodes in large-scale graphs. This task is challenging mainly due to 1) the intertwined nature of the node features and the graph structure calls for the feature condensation solver to be structure-aware; and 2) the difficulty of keeping useful information in the condensed graph. To address these challenges, we propose a novel framework {\tinygraph}, to condense features and nodes simultaneously in graphs. Specifically, we cast the problem as matching the gradients of GNN weights trained on the condensed graph and the gradients obtained from training over the original graph, where the feature condensation is achieved by a trainable function. The condensed graph obtained by minimizing the matching loss along the training trajectory can henceforth retain critical information in the original graph. Extensive experiments were carried out to demonstrate the effectiveness of the proposed {\tinygraph}. For example, a GNN trained with {\tinygraph} retains $98.5\%$ and $97.5\%$ of the original test accuracy on the \cora and \citeseer datasets, respectively, while significantly reducing the number of nodes by $97.4\%$ and $98.2\%$, and the number of features by $90.0\%$ on both datasets.
\end{abstract}

\begin{IEEEkeywords}
Graph Neural Networks, Data-efficient Learning, Graph Condensation.
\end{IEEEkeywords}

\section{Introduction}\label{sec:intro}
\IEEEPARstart{G}raphs have been extensively employed in modeling structured or relational real-world data~\citep{zhou2018graph,liu2023error,zhang2022contrastive}, encompassing various domains such as social network analysis~\citep{hamilton2017inductive,han2022geometric}, recommendation systems~\citep{ying2018graph,battaglia2018relational}, drug discovery~\citep{duvenaud2015convolutional,bongini2021molecular,xu2024llm}, transportation forecasting~\citep{xie2022explaining}, and epidemiology~\citep{liu2024review}.
Despite the success of GNNs in capturing abounding information in graph data~\citep{wu2019comprehensive-survey, han2022g}, training GNNs on real-world graphs containing large numbers of nodes and edges can be costly in terms of computational resources and time due to the complex sparse multiplication operations involved in training~\citep{zhang2021graph,joshi2022representation,hanmlpinit}. This problem is further compounded by the high-dimensional features that are common in graph data. More remarkably, the storage and time demands are magnified under the setting of automated machine learning~\citep{zhang2021automated,cai2022multimodal}, e.g., neural architecture search and hyperparameter optimization, where the GNN models need to be retrained for multiple times. One feasible idea to address these challenges is to condense the graph into a small graph so that it can save storage while facilitating the GNN training.
 \begin{figure}[t]
	\centering
    \includegraphics[width=1.0\linewidth]{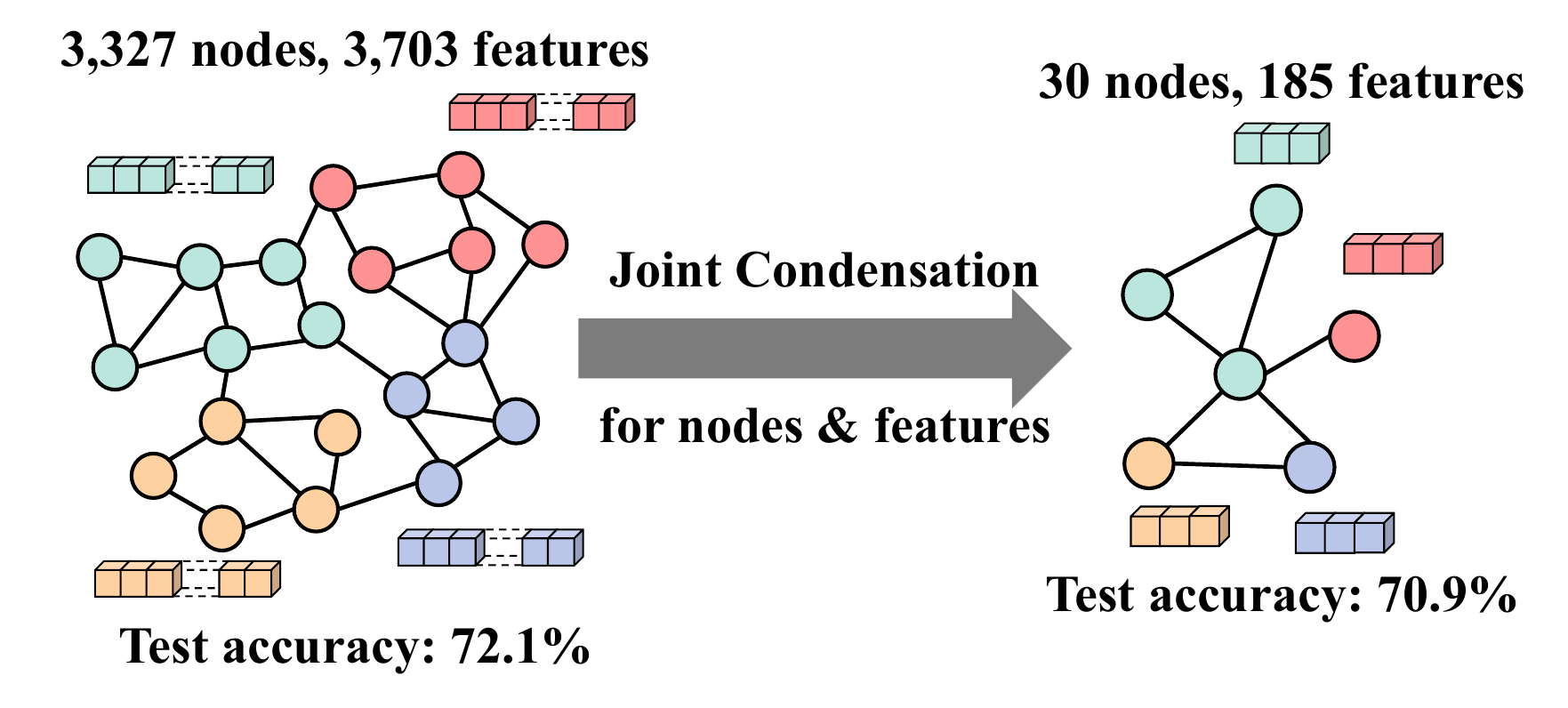}
\vspace{-1.8em}
	\caption{The goal of our proposed {\tinygraph}, which learn a condensed graph with much smaller node and feature sizes from a large graph. For example, {\tinygraph} is able to condense \citeseer with a data reduction of $90.0\%$ on features, $98.2\%$ on node size, and $99.7\%$ in storage.} \label{fig:omega}
 \vspace{-1.5em}
\end{figure}

Recent studies on graph condensation mainly focus on reducing the number of nodes in large-scale graphs~\citep{jin2021graph,gong2024gc,hashemi2024comprehensive}. 
An early work~\citep{zhao2020dataset} proposes a data condensation algorithm to reduce the training samples of image datasets. A more recent work~\citep{jin2021graph} generalizes the data condensation problem to the graph domain and designs a graph condensation framework that aims to compress the node size in a large graph.
Furthermore, a one-step gradient matching strategy has been proposed to improve the efficiency of graph condensation~\citep{jin2022condensing}.
However, existing graph condensation frameworks solely condense node size, which {may be insufficient} and still require large computation resources and storage when training {over graphs where the nodal features are high-dimensional.}
For example,
the number of features is $41$ times that of the number of {nodes} in \cora, $123$ times in \citeseer, $11.2$ times in \flickr, and $7.8$ times in \reddit. Such high dimensional features still lead to large weight matrices in GNN training. Hence only reducing the node size does not thoroughly resolve the issue of the high computational cost of GNN training.
As such, there is a crucial demand to effectively address the challenges posed by high-dimensional features in large-scale graph datasets.

In essence, to develop a structure-aware joint condensation framework for graph data, we are faced with two main challenges:
\textit{1) how to learn a tiny graph with fewer \textbf{nodes} and \textbf{features} that are still informative, so that the GNN trained on the tiny graph can {achieve comparable accuracy with that} trained on the original graph, and 
2) how to formulate the optimization problem and update {rules} for the joint condensation task that efficiently learns the condensed graph structure and features.} 
The first challenge {centers} around what needs to be learned and preserved from the original graph data while learning the condensed graph. 
One simple idea would be conducting the node and feature condensation sequentially by applying a two-stage framework, where one can carry out graph-condensation algorithms after dimensionality reduction is done on the nodal features. However, such a framework will likely lose critical information as structure-agnostic dimensionality reduction does not use the structural information often useful for graph-based learning. In addition, such a two-stage algorithm may result in error propagation if too much information is lost in the first stage.
The second challenge lies in how to circumvent the high computational complexity of multi-level optimization involved in joint condensation, as the learning procedure often requires updating the optimized parameters for GNN trained on the condensed graph as well as the trainable condensed graph in an iterative manner.

To this end, we propose a novel \textit{joint graph condensation} framework as illustrated in \cref{fig:omega}.
For the first challenge, we propose a unified framework {\tinygraph}, to simultaneously condense nodes and features. {\tinygraph} employs a gradient matching technique to {enforce the gradients} of the condensed graph to be as close as possible to the original graph along the training trajectory, and thus the learned condensed graph can be informative to train an accurate GNN. For the second challenge, {\tinygraph} does not rely on solving an optimization of the inner problem first and then an outer problem but only solves a gradient matching loss minimization problem for the learning task.
The goal of {\tinygraph} is to simultaneously condense nodes and features in a large graph while retaining {useful information} in the condensed graph. 
It synchronizes the GNN training {trajectories} of the two graphs through an optimization of a matching loss. This approach ensures that the condensed graph is consistent with the original graph and preserves the relationships between nodes.
This study makes the following major \textbf{contributions}:
\begin{itemize}[leftmargin=0.4cm, itemindent=0.0cm, itemsep=0.0cm, topsep=0.0cm]
    \item {We address the challenge that arises when training GNNs on large-scale graphs with high-dimensional features, which persists even in existing graph condensation methods.}
    \item 
    Instead of a two-stage technique, we propose a unified framework that condenses both nodes and features simultaneously. Specifically, a structure-aware feature condensation framework is designed that can efficiently take the graph structure into consideration. 
    \item {\tinygraph} is able 
    to achieve remarkable test accuracy of $98.5\%$, $97.5\%$, $98.7\%$, $94.9\%$, and $86.5\%$, meanwhile significantly reducing the node size by more than $97.4\%$ and the corresponding feature sizes by $90\%$, $90\%$, $80\%$, $70\%$, and $70\%$, on \cora, \citeseer, \flickr, \reddit, and \arxiv, respectively.
\end{itemize}

\section{Problem Formulation}
\noindent\textbf{Notations.} Given a graph dataset $\mathcal{T}=\{{\bf A}, {\bf X}, {\bf Y}\}$, {where ${\bf A}\in \mathbb{R}^{N\times N}$ is the adjacency matrix, $N$ is the number of nodes, ${\bf X}\in{\mathbb{R}^{N\times D}}$ is the node feature matrix and ${\bf Y}\in\{0,\ldots, C-1\}^N$ denotes the node labels over $C$ classes},  $\text{GNN}_{\boldsymbol{\theta}}$ represents a GNN model parameterized by $\boldsymbol{\theta}$, $\mathcal{L}$ denote the node classification loss (i.e., cross-entropy loss) that measures the discrepancy between predictions and the ground truth.

\noindent\textbf{Goal.} We aim to learn a tiny, condensed graph $\mathcal{S}=\{\hat{\bf {A}}, \hat{\bf {X}},\hat{\bf {Y}}\}$ with $\hat{\bf {A}}\in\mathbb{R}^{n\times n}$, $\hat{\bf {X}}\in\mathbb{R}^{n\times d}$, $\hat{\bf {Y}}\in\{0,\ldots, C-1\}^{n}$, {with $n<<N$, and $d<<D$}, such that a GNN trained on $\mathcal{S}$ can achieve comparable performance to one trained on the original graph $\mathcal{T}$. 
It is necessary for the larger graph to have the same feature dimension as $\mathcal{S}$ to match the input dimension of GNN.

\noindent\textbf{Problem Formulation.} To achieve this goal, we introduce a trainable feature condensation function $f_{\bf{\Phi}}(\cdot): \mathbb{R}^D\rightarrow \mathbb{R}^d$, parameterized on ${\bf{\Phi}}$. By projecting the original feature matrix ${\bf X}$ through this function, we obtain the feature-condensed graph $\mathcal{\tilde{T}}=({\bf A}, f_{\bf{\Phi}}({\bf X}), {\bf Y})$, which retains the key information from the full graph $\mathcal{T}$. This allows our proposed framework to establish an association between the original graph $\mathcal{T}$ and the condensed graph $\mathcal{S}$ through this function $f_{\bf{\Phi}}(\cdot)$. To ensure comparable performance achieved by training on the condensed graph and the original graph, we draw inspiration from prior studies~\citep{wang2018dataset, sucholutsky2019soft, bohdal2020flexible, such2020generative}, that models the parameters $\boldsymbol{\theta}_{\mathcal{S}}$ as a function of the synthetic data $\mathcal{S}$. Henceforth, we approach the joint condensation problem by training a GNN on the trainable condensed graph $\mathcal{S}$, based on which the feature condensation function parameterized by ${\bf \Phi}^{*}$ and condensed graph $\mathcal{S}^{*}$ is obtained by minimizing the loss. The objective of the joint condensation problem can be formularized as follows:
\begin{equation}
\begin{split}
\mathcal{S}^{*}=&\underset{\mathcal{S}}{\arg \min } \; \mathcal{L}(\text{GNN}_{\boldsymbol{\theta}^{*}_{\mathcal{S}}}({\bf A},f_{\bf{\Phi}^{*}}(\bbX)), {\bf Y}),
\\
&\text {s.t.} \ {\bf{\Phi}^{*}}=\underset{\bf{\Phi}}{\arg \min } \; \mathcal{L}(\text{GNN}_{\boldsymbol{\theta}^{*}_{\mathcal{S}}}({\bf A},f_{\bf{\Phi}}(\bbX)), {\bf Y}),\\
&\boldsymbol{\theta}^{*}_{\mathcal{S}}=\underset{\boldsymbol{\theta}_{\mathcal{S}}}{\arg \min } \; \mathcal{L}(\text{GNN}_{\boldsymbol{\theta}_{\mathcal{S}}}(\hat{\bf {A}}, \hat{\bf {X}}), \hat{\bf {Y}}).
\label{eq:bi_level_2}
\end{split}
\end{equation}

Note that the optimization in~\cref{eq:bi_level_2} necessitates solving a nested-loop optimization problem. Specifically, this involves iteratively updating $\boldsymbol{\theta}_{\mathcal{S}}$ while performing computationally expensive computations of the gradient of the trainable condensed graph $\mathcal{S}$, i.e., $\mathcal{L}^{\mathcal{S}}$, which is obtained by calculating the discrepancies between the model predictions and the ground truth for $\mathcal{S}$. For calculating $\mathcal{L}$, We employ the cross-entropy loss. These computations are carried out over multiple updates within each iteration.

However, this procedure encounters scalability issues when applied to large graphs with a substantial number of optimization steps in the inner loop. The computational burden imposed by these iterations becomes increasingly prohibitive as the graph size grows. To address this limitation, we propose an alternative algorithm in the subsequent section. Our approach follows the gradient matching strategy~\citep{liu2023fairgraph}, which aims to overcome the scalability challenges in nested-loop optimization and offers an efficient solution to the joint condensation.

\section{Proposed Algorithm}
In this section, we introduce a novel algorithm {\tinygraph} to solve the problem in~\cref{eq:bi_level_2}.
 \begin{figure*}[t]
	\centering
    \includegraphics[width=1.0\linewidth]{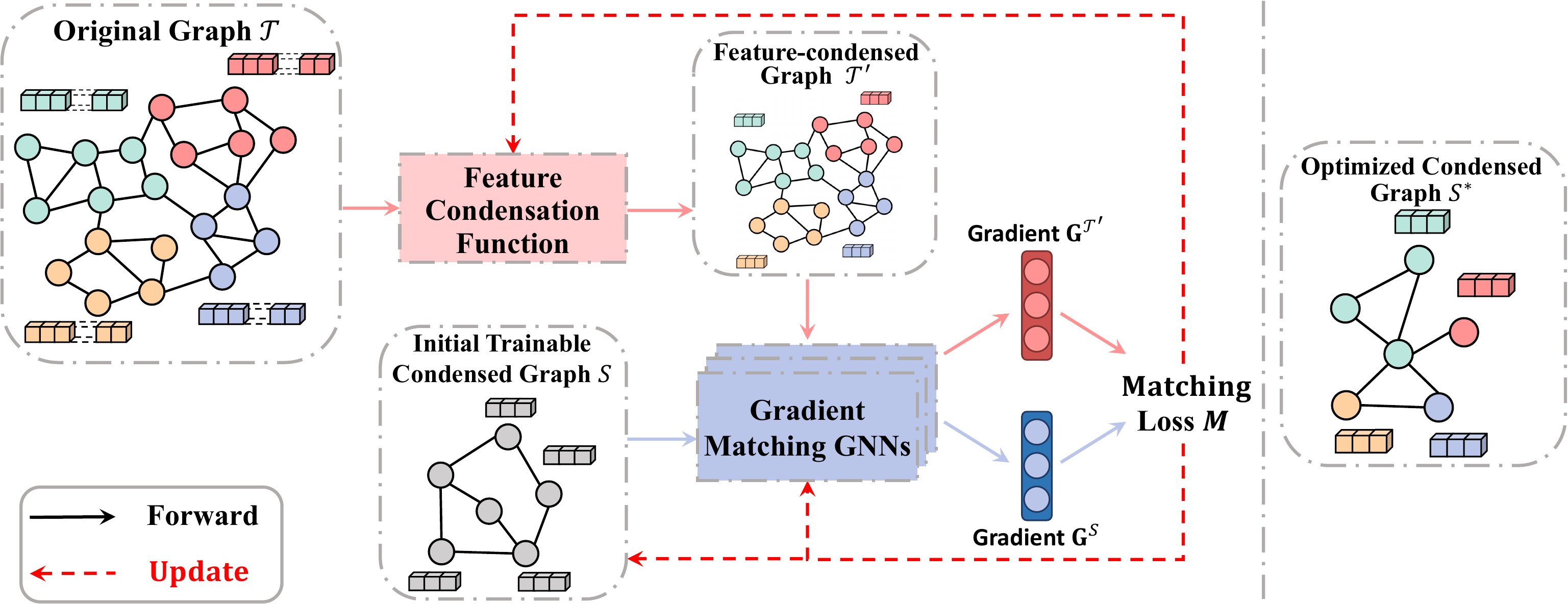}
     \vspace{-0.8em}
	\caption{ An overview of the {\tinygraph} framework for addressing the joint condensation problem via gradient matching. The objective is to learn a condensed `tiny' graph that can be used to train a GNN, achieving performance comparable to training on the original graph. Both the feature condensation function and the gradient matching GNN are trainable.
 } 
 \label{fig:framework}
 \vspace{-1.2em}
\end{figure*}
\subsection{Structure-aware feature condensation}
 
Graph data presents unique challenges for traditional dimensionality reduction (DR) methods. Principal Component Analysis (PCA)~\citep{abdi2010principal}, or Independent Component Analysis (ICA)~\citep{comon1994independent} are two typical and widely used DR methods. Others include linear and nonlinear techniques, unsupervised and supervised methods, and matrix factorization and manifold learning approaches. They have been shown promising results in abstracting primary features in various research disciplines, such as on image data~\citep{luo2020dimensionality,zhou2019discriminative}, text data~\citep{kowsari2019text,zebari2020comprehensive}, as well as biological or health data~\citep{becht2019dimensionality}. However, their applications are limited to Euclidean data where samples are independent and thus are not well-suited for graph data where nodes are interlaced with others based on graph structure. Therefore, grafting these methods for graph data is not feasible, and it is necessary to design a structure-aware algorithm for high-dimensional graph data.
To better capture the graph structure in the condensation process, we adopt a graph attention function~\citep{velivckovic2017graph} as the feature condensation function, which contains multiple graph attention networks (GAT) layers. GAT is an attention-based GNN model that leverages attention mechanisms to assign precise weights during the aggregation of neighboring node information pertaining to a given target node. By incorporating the fine-grained weight assignment capability of GAT, the proposed approach aims to improve the accuracy and comprehensiveness of the graph condensation process.
 
\noindent\textbf{Graph Attention Networks as Feature Condensation Function.}
The graph attentive network in the condensation framework is denoted as $\text{GAT}_{{\bf{\Phi}}_{t}}$. It is parameterized by ${\bf{\Phi}}_{t}$ and maps the original features $\bbX$ to the trainable condensed features $\tilde{\bf X}_{t}=\text{GAT}_{{\bf{\Phi}}_{t}}(\mathbf{X})$. 
The graph that contains the trainable condensed features, is referred to as a trainable condensed graph denoted by $\mathcal{\tilde{T}}=({\bf A}, \tilde{\bf X}_t, {\bf Y})$.
During each training epoch $t$, the node representation of node $v_i$ is obtained. For brevity, the epoch index $t$ will be omitted in the subsequent notations pertaining to the calculation of the Graph Attentive Encoder~\citep{ding2021cross}. Within each layer of the GAE, denoted by $l$, node $v_i$ integrates the features of its neighboring nodes to acquire representations for the subsequent layer, $l+1$, through the following process:
$$
\mathbf{h}_{i}^{l+1}=\sigma\left(\sum_{j \in \mathcal{N}_i \cup v_i} \alpha_{ij}^{l} \mathbf{W h}_{j}^{l}\right),
$$
where $\mathbf{h}_i^{l+1}$ represents the representation of node $v_i$ in layer $l+1$, $\mathcal{N}i$ denotes the set of neighboring nodes of $v_i$, $\mathbf{W}^{l}$ represents a weight matrix associated with layer $l$, and $\sigma$ denotes a nonlinear activation function (e.g., ReLU), and $\alpha_{ij}^{l}$ is an attention coefficient that determines the importance of neighboring node $v_j$ to node $v_i$ in layer $l$, which can be computed as:
$$
\alpha_{ij}^{l}=\frac{\exp \left(\sigma\left(\mathbf{a}^{\top}\left[\mathbf{W} \mathbf{h}_{i}^{l} \oplus \mathbf{W} \mathbf{h}_{j}^{l}\right]\right)\right)}{\sum_{k \in \mathcal{N}_i \cup v_i} \exp \left(\sigma\left(\mathbf{a}^\top\left[\mathbf{W}_{t} \mathbf{h}_{t}^{l} \oplus \mathbf{W} \mathbf{h}_{k}^{l}\right]\right)\right)},
$$
where $\mathbf{W}$ and $\mathbf{W}_{t}$ represent weight matrices, and $\mathbf{h}_{i}^{l}$ and $\mathbf{h}_{j}^{l}$ denote the hidden states of nodes $v_i$ and $v_j$ in layer $l$, respectively. The symbol $\oplus$ denotes the concatenation operation. The term $\mathbf{a}$ represents a trainable parameter vector that allows the model to learn the importance of different node relationships.
Furthermore, in order to enhance the expressive power of our model, we incorporate multiple GAT layers into GAE. The computation of the latent representation, denoted as $\mathbf{z}_i$, for each node $i$ is performed as follows:
$$
\begin{aligned}
& \hat{\mathbf{x}}_i=\sigma\left(\sum_{j \in \mathcal{N}_i \cup v_i} \alpha_{i j}^{L} \mathbf{W}^{L} \mathbf{h}_{j}^{L-1}\right),
\end{aligned}
$$
where $\hat{\mathbf{x}}_i\in \mathbb{R}^d$ ($d<<D$) is the trainable condensed feature of node $i$, and $\mathbf{h}_{j}^{L-1}$ is the hidden representation of node $j$ at layer $L-1$. We initialize the first layer by setting $\mathbf{h}_{j}^{1}=\mathbf{x}_j$, which is the nodal feature of node $i$.
In each training epoch, we encode the original feature matrix $\mathbf{X}$ using the GAE. By employing GAE, the proposed condensation framework is capable of learning the {condensed} feature while considering the underlying graph structure.
By incorporating multiple GAT layers, {\tinygraph} leverages the interplay between nodal attributes and the graph topology, our model captures the intricate non-linear relationships present in the graph. This enhanced representation learning capability is essential for addressing the complexities of real-world graph data and facilitating downstream tasks that rely on learning an accurate condensed graph.

\subsection{Gradient Matching}
Note that the objective of this study is to learn a condensed graph, denoted as $\mathcal{S}$, that allows the trained model to converge to a solution similar to that of an optimized parameter space represented by ${\mathcal{\tilde{T}}}$. To this end, an intuitive optimization approach is to align the optimal GNN parameters trained from the trainable feature-condensed graph and the {trainable condesed} graph, denoted as $\boldsymbol{\theta}_{\mathcal{\tilde{T}}}^{*}$ and $\boldsymbol{\theta}_{\mathcal{{S}}}^{*}$, respectively, In this way, the resulting optimized parameters converge to a comparable solution:
\begin{equation}
\begin{split}
&\min_{\mathcal{S},{\bf \Phi}}
M\left(\boldsymbol{\theta}^{*}_{\mathcal{S}}, \boldsymbol{\theta}^{*}_{\mathcal{\tilde{T}}}\right) 
\\
&\quad\text{ s.t. }\boldsymbol{\theta}^{*}_{\mathcal{S}}=\underset{\boldsymbol{\theta}_{\mathcal{S}}^{t}}{\arg \min }\mathcal{L}\left(\text{GNN}_{\boldsymbol{\theta}_{\mathcal{S}}^{t}}(\hat{\bf {A}}, \hat{\bf {X}}), \hat{\bf {Y}}\right),
\\
&\quad\text{and }\boldsymbol{\theta}^{*}_{\mathcal{\tilde{T}}}=\underset{\boldsymbol{\theta}_{\mathcal{\tilde{T}}}^{t}}{\arg \min }\mathcal{L}\left(\text{GNN}_{\boldsymbol{\theta}_{\mathcal{\tilde{T}}}^{t}}({\bf A},\tilde{\bf X}), {\bf Y}\right),
\label{eq:one_model}
\end{split}
\end{equation}
where the objective is to minimize the matching loss $M(\cdot,\cdot)$ with respect to the optimal parameters $\boldsymbol{\theta}_{\mathcal{S}}^{*}$ and $\boldsymbol{\theta}_{\mathcal{\tilde{T}}}^{*}$, which are associated with the condensed graph and the model $\text{GNN}_{\boldsymbol{\theta}_{\mathcal{\tilde{T}}}^{t}}$ respectively. The optimization of \cref{eq:one_model} aims to obtain a trainable condensed graph $\mathcal{S}$ for the GNN model that is parameterized by $\boldsymbol{\theta}^{*}_{\mathcal{\tilde{T}}}$.
However, the effectiveness of $\boldsymbol{\theta}_{\mathcal{\tilde{T}}}^{*}$ is influenced by its initial value $\boldsymbol{\theta}_{\mathcal{\tilde{T}}}^{0}$. Consequently, the objective function in \cref{eq:one_model} optimizes over a single model initialized with the initial weights $\boldsymbol{\theta}_{\mathcal{\tilde{T}}}^{0}$. Our objective is to ensure that $\boldsymbol{\theta}_{\mathcal{S}}^{*}$ not only closely approximates the final value of $\boldsymbol{\theta}^{*}_{\mathcal{\tilde{T}}}$, but also follows a similar trajectory to the parameter values $\boldsymbol{\theta}_{\mathcal{\tilde{T}}}^{t}$ at time $t$ throughout the optimization process.

\noindent\textbf{Curriculum Gradient Matching.}
In order to enhance the learning process and improve optimization outcomes, we adopt \textit{curriculum gradient matching}~\citep{wang2018dataset}. This technique aims to align the training trajectories of $\mathcal{\tilde{T}}$ and $\mathcal{S}$, throughout each epoch of the training process, achieved by quantifying the disparity between their respective gradients.
By employing curriculum gradient matching, the problem in \cref{eq:one_model} can be reformulated as:
\begin{equation}
\begin{split}
&\min _{\mathcal{S}} \mathrm{E}_{\boldsymbol{\theta}^{0}\sim Q_{\boldsymbol{\theta}^0}}\left[\sum_{t=0}^{T-1} 
M\left(\boldsymbol{\theta}^{t}_{\mathcal{S}}, \boldsymbol{\theta}^{t}_{\mathcal{\tilde{T}}}\right)\right] 
\\
\ &\text{ with } \
\boldsymbol{\theta}^{t+1}_{\mathcal{S}} \!=\! \boldsymbol{\theta}^{t}_{\mathcal{S}}\!-\!\eta \nabla_{\boldsymbol{\theta}_{\mathcal{S}}^{t}} \mathcal{L}\!\left(\!\text{GNN}_{\boldsymbol{\theta}_{\mathcal{S}}^{t}}(\hat{\bf {A}},\hat{\bf {X}}), \hat{\bf {Y}}\!\right), \\  
&\text{ and } \ \boldsymbol{\theta}^{t+1}_{\mathcal{\tilde{T}}}\!=\! \boldsymbol{\theta}_{\mathcal{\tilde{T}}}^{t}\!-\!\eta \nabla_{\boldsymbol{\theta}_{\mathcal{\tilde{T}}}^{t}}
\mathcal{L}\left(\text{GNN}_{\boldsymbol{\theta}_{\mathcal{\tilde{T}}}^{t}}({\bf A},\tilde{\bf X}), {\bf Y}\right),\label{objective1}
\end{split}
\end{equation}
where  $\text{GNN}_{\boldsymbol{\theta}_{\mathcal{\tilde{T}}}^{t}}$ denotes the GNN model parameterized with $\boldsymbol{\theta}_{\mathcal{\tilde{T}}}^{t}$, 
$\eta$ is the learning rate for the gradient descent. 
This approximation method eliminates the need for computationally expensive unrolling of the recursive computation graph over the previous parameters, denoted as ${{\bf \theta}^{0}, \dots, {\bf \theta}^{t-1}}$. Consequently, the optimization technique demonstrates significantly enhanced speed, 
improved memory efficiency, and the ability to scale up to the joint condensation. By minimizing $M$ to a near-zero value, we aim to obtain the optimal $\boldsymbol{\theta}_{\mathcal{S}}^{*}=\boldsymbol{\theta}_{\mathcal{S}}^{t+1}$, which is parameterized by $\mathcal{S}$, by emulating the training of $\boldsymbol{\theta}_{\mathcal{\tilde{T}}}^{t+1}$. 

\noindent\textbf{Gradient Matching Loss.}
Based on the optimization approach described earlier, we propose the utilization of a gradient-matching loss to update the trainable parameters.
In this context, we have two losses: $\mathcal{L}^{\mathcal{\tilde{T}}}$ and $\mathcal{L}^{\mathcal{S}}$, which represent the discrepancies between the model predictions and the ground truth for the transformed original graph and the {trainable condensed} graph, respectively. For calculating $\mathcal{L}$, We employ the cross-entropy loss.
To update the weights of GNN, we calculate the gradients of the losses with respect to the GNN weights. Specifically, we denote these gradients as $\mathbf{G}^{\mathcal{\tilde{T}}}$ and $\mathbf{G}^{\mathcal{S}}$, which are obtained through the respective expressions:
$\mathbf{G}^{\mathcal{\tilde{T}}}=\nabla_{\boldsymbol{\theta}}\mathcal{L}(\text{GNN}_{{\boldsymbol{\theta}}}({\bf A},f_{\bf{\Phi}}(\bbX)), {\bf Y})$ and $\mathbf{G}^{\mathcal{{S}}}=\nabla_{\boldsymbol{\theta}}\mathcal{L}\left(\text{GNN}_{\boldsymbol{\theta}}(\hat{\bf {A}}, \hat{\bf {X}}), \hat{\bf {Y}}\right)$.

The gradient matching loss $M$ quantifies the dissimilarity between the gradients of the source network and the target network training.
To calculate the gradient matching loss $M$, we consider each individual class $c$ separately.
For a given class $c$, the gradient distance is determined by computing the Cosine Distance between the corresponding columns of the gradient matrices, i.e., $\mathbf{G}^{\mathcal{\tilde{T}}^{c}}$ and $\mathbf{G}^{\mathcal{{S}}^{c}}$. These matrices represent the gradients of the sampled graphs $\mathcal{\tilde{T}}^{c}$ and $\mathcal{{S}}^{c}$ of a class $c$, derived from the original graph and condensed graph, respectively.
This Cosine Distance captures the disparity in the way class-specific information is represented in the two graphs. By assessing the Cosine Distance between the gradient columns, we quantify the dissimilarity between the gradient directions associated with class $c$ in the original and condensed graphs. 
By summing the gradient distances across all classes, the gradient matching loss $M$ provides a comprehensive assessment of the overall dissimilarity between the networks.
Mathematically, the gradient matching loss $M$ can be formulated as follows:
\begin{equation}
\begin{split}
M(\mathbf{G}^{\mathcal{\tilde{T}}}, \mathbf{G}^{\mathcal{{S}}})&=\sum_{c=0}^{C-1}M^{c}\\
\text{with}\ M^{c}&
=\sum_{p=1}^{P}\sum_{h=1}^{H }\!\!\left(1-\frac{
\mathbf{G}^{\mathcal{\tilde{T}}^{c}}_{h,p}
\cdot 
\mathbf{G}^{\mathcal{{S}}^{c}}_{h,p}}
{\left\|\mathbf{G}^{\mathcal{\tilde{T}}^{c}}_{h,p}\right\|\left\|\mathbf{G}^{\mathcal{{S}}^{c}}_{h,p}\right\|}\right),
\label{eq:matching-loss}
\end{split}
\end{equation}

where $\mathbf{G}^{\mathcal{\tilde{T}}^{c}}_{h,p}$ and $ \mathbf{G}^{\mathcal{S}^{c}}_{h,p}$ are the $h$-th column vectors of the gradient matrices for the class $c$, at layer $p$. 
In the subsequent subsection, we delve deeper into the practical implementation and utilization of the gradient-matching process, discussing how it is integrated into our framework to enhance the overall condensation procedure and optimize the trainable parameters.

\subsection{Model Optimization}
The standard large-batch optimization process poses challenges for reconstruction tasks and necessitates substantial memory usage~\citep{zhu19deep}. To address this, we adopt a strategy where a mini-batch of graph data is sampled at each layer of GNN. Subsequently, we compute the gradient-matching loss for each class independently.
Specifically, to handle a specific class $c$, {\tinygraph} selects a batch of nodes and their corresponding neighbors from the transformed original graph $\mathcal{\tilde{T}}$, denoted as $\mathcal{\tilde{T}}^{c}\sim\mathcal{\tilde{T}}$. Likewise, {\tinygraph} samples a batch of condensed nodes from the condensed graph $\mathcal{S}={\hat{\bf {A}}, \hat{\bf {X}},\hat{\bf {Y}}}$ and incorporates all of their neighbors, denoted as $\mathcal{S}^{c} \sim\mathcal{S}$.

Despite these measures, the training process remains challenging primarily due to the significant complexity introduced by learning graph structure. This complexity stems from factors such as intricate relationships between nodes, intricate connectivity patterns, and the need to capture graph-specific features. Hence, in the following, we will further navigate these challenges and improve the effectiveness of joint condensation.
 
\noindent\textbf{Parameterizing Graph Structure.}
Treating $\hat{\bf{A}}$ and $\hat{\bf{X}}$ as free parameters will lead to overfitting caused by learning a large number of parameters in the order of $O({n}^2)$.
In order to alleviate the computational complexity associated with optimizing the quadratic model of $\hat{\bf{A}}\in\mathbb{R}^{n\times n}$, we model $\hat{\bf{A}}$ as a function $\bf{\Psi}$, denoted by $g_{\bf{\Psi}}(\cdot)$, which is parameterized by $\bf{\Psi}$~\citep{jin2021graph}, i.e.:
\begin{equation}
\begin{aligned}  
        \hat{\bf A}_{ij}=g_{\bf{\Psi}} (\hat{\bf {X}}_{ij}) = \sigma\left(\frac{{k_{\bf{\Psi}}(\mathbf{z})}+{k_{\bf{\Psi}}(\mathbf{z}')}}{2}\right),
\label{eq:adj}
\end{aligned}
\end{equation}
where both vectors $\mathbf{z}:=[\hat{\mathbf{x}}^{\top}_i;\hat{\mathbf{x}}^{\top}_j]$ and $\mathbf{z}':=[\hat{\mathbf{x}}^{\top}_j;\hat{\mathbf{x}}^{\top}_i]$ are elements of $\mathbb{R}^{1\times 2d}$. The function $k_{\boldsymbol{\Psi}}(\cdot)$ denotes a multi-layer neural network (MLP) parameterized by $\boldsymbol{\Psi}$, and $\sigma(\cdot)$ represents a sigmoid activation function.
This method offers the scalability to expand the condensed graph by incorporating additional synthetic nodes derived from the real graph. In this expansion process, the trained $g_{\bf \Psi}(\cdot)$ can be effectively utilized to infer the connections of the newly added synthetic nodes. As a result, we only need to focus on learning the distinctive features of these new nodes, while leveraging the existing graph structure and connections inferred by $g_{\bf \Psi}(\cdot)$.

Building upon the aforementioned elucidation, we articulate the ultimate objective of the proposed framework. This objective is formulated based on an empirical observation: the proximity between $\boldsymbol{\theta}^{t}_{\mathcal{S}}$ and $\boldsymbol{\theta}^{t}_{\mathcal{\tilde{T}}}$ is typically small. Consequently, we couple them and replace them with $\boldsymbol{\theta}^t$, denoting the GNN weights trained on $\mathcal{S}$ at time $t$. In light of this, we can simplify the objective presented in \cref{objective1} by treating it as a gradient-matching process. This process can be represented in the following manner:
\begin{equation}
\begin{split}
&\min_{\hat{\bf {X}}, \bf{\Phi}, \bf{\Psi}} \!\!\mathrm{E}_{\boldsymbol{\theta}^{0}\sim Q_{\boldsymbol{\theta}^0}}\!\biggl[\sum_{t=0}^{T-1}\!M\!\Bigg(\!\!\nabla_{\boldsymbol{\theta}^{t}}\mathcal{L}\left(\!\text{GNN}_{\boldsymbol{\theta}^{t}}\!(g_{\bf{\Psi}}(\hat{\bf {X}}),\hat{\bf {X}}), \hat{\bf {Y}}\!\right),\\
&\quad\quad\quad\quad\quad\quad\quad\quad\nabla_{\boldsymbol{\theta}^{t}}\mathcal{L}\left( \text{GNN}_{\boldsymbol{\theta}^{t}}({\bf A},f_{{\bf{\Phi}}_{t}}({\bf{X}})), {\bf Y}\right)\Bigg)\biggl].
\label{eq:final}
\end{split}
\end{equation} 
The loss function $M$ combines the gradients of $\mathcal{L}$ with respect to the parameters $\boldsymbol{\theta}^t$ for both input cases. The minimization of this loss function drives the optimization process, influencing the updates made to the variables $\hat{\bf {X}}$, $\boldsymbol{\Phi}$, and $\bf{\Psi}$. {\cref{fig:framework}} provides an overview of our proposed framework and depicts the optimization process visually.

\noindent\textbf{Alternative Optimization Strategy.}
Optimizing $\hat{\bf {X}}$, $\bf{\Psi}$, and $\bf{\Phi}$ simultaneously can be a challenging task due to their interdependence. To overcome this challenge, we employ an alternative optimization strategy in our research. Our approach aims to iteratively update the parameters $\bf{\Phi}$, ${\bf{\Psi}}$, and $\hat{\bf {X}}$ in distinct time periods.
At each epoch, our method begins by updating the feature condensation function $\bf{\Phi}$. Following this, for the first $t_1$ training epochs, we focus on updating ${\bf{\Psi}}$. Then we proceed to update the condensed features $\hat{\bf {X}}$ for the subsequent $t_2$ epochs. 

Importantly, the parameters are updated asynchronously at different epochs, reflecting their interdependence. This asynchronous updating scheme allows us to leverage the information and progress made in the previous phases, leading to more effective optimization.
We iterate this process until a predefined stopping condition is met, such as reaching a convergence criterion or completing a fixed number of epochs:
\begin{equation}
\begin{array}{c}
{\bf \Phi}_{t+1} = {\bf \Phi}_{t} -\eta_1 \nabla_{{\bf \Phi}} M \  \text{(every epoch)},\\
{\bf{\Psi}}_{t+1} - {\bf{\Psi}}_{t} -\eta_2 \nabla_{\bf{\Psi}} M \  \text{($t_1$ epochs)},\\
\hat{\bf {X}}_{t+1} = {\hat{\bf {X}}_{t}}-\eta_3 \nabla_{\hat{\bf {X}}} M  \ \text{($t_2$ epochs)}.\\
\end{array}
\end{equation}

This iterative optimization strategy ensures that each parameter is updated strategically, accounting for their mutual influence and optimizing the overall performance of the model.

\noindent\textbf{Model Initialization.}
For the initialization of trainable parameters in the joint condensation, we centralize original features by $\mathbf{X}'=\mathbf{X}({\mathbf I}_D- \frac{1}{D}  {\mathbf 1}_{D}{\mathbf 1}_{D}^\top )$.
To simplify the joint condensation, we fix the node labels ${\bf \hat{Y}}$ while keeping the class distribution as original labels ${\bf Y}$. For the initialization of the trainable condensed graph, we use graph sampling. We first sample a subset of nodes from the trainable condensed graph $\mathcal{\tilde{T}}=({{\bf A}, {\tilde{\bf X}}, {\bf Y}})$, where $\tilde{\bf X}=f_{\bf{\Phi}}({\bf X})\in{\mathbb{R}^{N\times d}}$ is the trainable condensed feature. The number of sampled nodes from each class is set to preserve the distribution of the labels.  
Learning all four variables, namely $\hat{\mathbf{X}}$, $\hat{\mathbf{Y}}$, $\boldsymbol{\Phi}$, and $\boldsymbol{\Psi}$, poses significant challenges. To simplify the problem, we fix the node labels $\hat{\mathbf{Y}}$ after initialization, and the feature vectors corresponding to ${\bf \hat{Y}}$ are used to initialize $\hat{\bbX}\in\mathbf{R}^{n\times d }$.
\begin{algorithm}[t]
\SetAlgoVlined
\small
\textbf{Input:} Training data
$\mathcal{T}\!=\!({\bf A}, {\bf X}, {\bf Y})$, pre-defined labels $\hat{\bf {Y}}$. \\
Initialize $\hat{\bf {X}}$ of dimension $d$ by randomly selecting node features from each class.\\
\For{$k=0,\ldots, K-1$}{
  Initialize $\boldsymbol{\theta}^{0}\sim Q_{\boldsymbol{\theta}^0}$\\
  \For{$t=0,\ldots,T-1$}{
  $M = 0$ \\
Compute $\tilde{\bf X}_t=f_{{\bf \Phi}_t}({\bf X})$ and to form $\mathcal{\tilde{T}}$\\
  \For{$c=0,\ldots,C-1$}{
  Compute $\hat{\bf {A}} = g_{{\bf \Psi}_t}(\hat{\bf {X}})$; then $\mathcal{S}=\{\hat{\bf {A}}, \hat{\bf {X}}, \hat{\bf {Y}}\}$ \\
  Sample 
  $(\tilde{\bf A}^{c},\tilde{\bf X}^{c}, {\bf Y}^{c})\sim \mathcal{\tilde{T}}$ and $(\hat{\bf {A}}^{c},\hat{\bf {X}}^{c}, \hat{\bf {Y}}^{c})\sim \mathcal{S}$ \ $\rhd$  detailed in Section 3.1 
  \\
    Compute $\nabla_{\boldsymbol{\theta}^t}\mathcal{L}^{\mathcal{\tilde{T}}}, \text{and} \ \nabla_{\boldsymbol{\theta}^t}\mathcal{L}^{\mathcal{S}}$
  \\
Obtain $M$ from \cref{eq:matching-loss}\\
}  
Update ${\bf \Phi}_{t+1} = {\bf \Phi}_{t} -\eta_1 \nabla_{{\bf \Phi}_{t}} M$

 \If{$t \% (t_1+t_2) < t_1$}{
 Update ${\bf \Psi}_{t+1} = {\bf \Psi}_{t} -\eta_2 \nabla_{{\bf \Psi}_{t}} M$\\
  }
 \Else{
Update $\hat{\bf {X}}_{t+1} = \hat{\bf {X}}_{t} -\eta_3 \nabla_{\hat{\bf {X}}_{t}} M$
}
 
\For{$j=0,\ldots,J-1$}{
Update $\boldsymbol{\theta}^{t+1} = \boldsymbol{\theta}^{t}-\eta \nabla_{\boldsymbol{\theta}^t} \mathcal{L}^{\mathcal{S}}$}
}}
 $\hat{\bf {A}}= g_{\bf \Psi}(\hat{\bf {X}})$\\
   {$\hat{\bf {A}}_{ij}= \hat{\bf {A}}_{ij}$ if ${\hat{\bf {A}}_{ij}} > \gamma$, otherwise $0$}\\
 \textbf{Return:} $\mathcal{S}=(\hat{\bf {A}}_{T-1}, \hat{\bf {X}}_{T-1}, \hat{\bf {Y}}_{T-1}),\text{and} \ {\bf {\Phi}}_{T-1}$\\
\caption{{\tinygraph} for Graph Condensation}\label{gcond_pca_algo1}
\end{algorithm}

 \noindent\textbf{Algorithm Implementation.}
In our algorithm, we follow a series of steps to implement the {\tinygraph} framework. We begin by initializing the GNN model parameter $\boldsymbol{\theta}^{0}$, which is sampled from a uniform distribution $Q_{\boldsymbol{\theta}^0}$ based on $\boldsymbol{\theta}^{0}$.
Next, we proceed with the sampling of node batches from the labeled training graph $\mathcal{T}$ and the condensed graph $\mathcal{S}$ for each class. These batches serve as input for the subsequent computations. Within each class, we calculate the gradient matching loss. The losses obtained from each class are summed up and then used to update specific parameters such as ${\bf{\Phi}}$, $\hat{\bf {X}}$, and ${\bf{\Psi}}$. 
Once the condensed graph parameters have been updated, the GNN parameters are updated for a specified number of epochs $t_{\boldsymbol{\theta}}$. During this phase, the GNN model is fine-tuned to improve its performance.
Finally, to obtain the final sparsified graph structure, we apply a filtering step. We discard edge weights that fall below a predetermined threshold $\gamma$. This filtering process helps in simplifying the graph representation and reducing unnecessary edges.
The detailed algorithm of {\tinygraph} is summarized in \cref{gcond_pca_algo1}.

\subsection{Discussion on the Differences from Related Studies}
In this subsection, we further discuss the novelty of the proposed {\tinygraph} compared with node condensation methods and two-stage condensation approaches.

\noindent\textbf{Comparison with Node Condensation Methods.}
Various recent works have focused on reducing the size of data while preserving essential information, such as data condensation methods~\citep{zhao2020dataset, jin2021graph, jin2022condensing, bai2021condensing} and graph compression methods~\citep{wang2020graph, spielman2011graph}. However, these methods overlook the high dimensionality of nodal features in real-world graphs. In contrast, our proposed method addresses both feature and node condensation, achieving significant storage reduction of $75.3\%$, $83.3\%$, $87.0\%$, $87.8\%$, and $51\%$ on \cora, \citeseer, \flickr, \reddit, and \arxiv, respectively, when compared to the existing node condensation framework {\method}.

\noindent\textbf{Two-stage Condensation Approaches.}
Two-stage condensation approaches directly apply dimensionality reduction methods~\citep{abdi2010principal, comon1994independent, van2008visualizing, balakrishnama1998linear, lee1999learning, vincent2008extracting, bank2020autoencoders, kingma2019introduction} to node condensation frameworks, which are insufficient due to the structure-agnostic nature in their feature condensation process. Compared to these approaches, {\tinygraph} offers two key advantages in feature condensation: 1) the feature and node condensation are learned through unified optimization, enabling structure awareness in feature condensation, and 2) {\tinygraph} utilizes a learnable projection function for feature condensation, offering greater flexibility than fixed projection matrices used in DR methods.

\section{Experiments}\label{sec:experiments}
In this section, we conduct experiments with five real-world graphs to evaluate {\tinygraph}. The main observations in experiments are highlighted as \Circled{\footnotesize \#} \textbf{boldface}.
\subsection{Experimental setup.}
\begin{table}[!t]
\setlength{\tabcolsep}{0.7pt}
\centering
\caption{Statistics and properties of five datasets. \cora, \citeseer, and \arxiv are transductive datasets. \flickr and \reddit are inductive datasets. $\texttt{C-Seer}$ denotes the \citeseer dataset.}
\vspace{-0.8em}
\begin{tabular}{l|c|c|c|c|c}
\toprule
{Dataset} & {\# Nodes} &{\# Feat.} & {\# Edges} & {\# Classes} &
{Train/Validation/Test} \\
\midrule 
\cora & $2,708$ & $1,433$ & $5,429$ & $7$ & $140$/$500$/$1,000$\\
\texttt{C-Seer} & $3,327$ & $3,703$ & $4,732$ & $6$ & $120$/$500$/$1,000$\\
\flickr & $89,250$ & $500$ & $899,756$ & $7$ & $44,625$/$22,312$/$22,313$\\
\reddit & $232,965$ & $602$ & $57,307,946$ & $210$ & $153,932$/$23,699$/$55,334$\\
\arxiv & $169,343$ &$128$ &$1,166,243$ & $40$ & $44,625/22,312/22,313$\\
\bottomrule
\end{tabular}
\label{tab:data}
\vspace{-2em}
\end{table}
\noindent\textbf{Datasets.} 
{We evaluate the efficacy of our proposed method over three transductive datasets, i.e., \cora, \citeseer~\citep{kipf2016semi}, and \arxiv~\citep{hu2020open}, as well as two inductive datasets, i.e., \flickr~\citep{graphsaint-iclr20} and \reddit~\citep{hamilton2017inductive}. For consistency and fair comparisons, we utilized all the datasets provided by PyTorch Geometric\citep{fey2019fast-pyg}, following the publicly available data splits, as widely adopted by previous studies~\citep{hamilton2017inductive}. Note that the employed experimental setup is outlined in~\citep{hamilton2017inductive}. We also present the detailed statistics of the used in \cref{tab:data}.

\noindent\textbf{Baselines.}
We conducted a comprehensive comparative analysis between our proposed method, {\tinygraph}, and seven baseline approaches, which are: (1) a model that utilizes the \textit{original} graph structure and node features without any condensation for training; (2) and (3) two unsupervised classical methods for \textit{density-based clustering}, which employ condensed graph structures and complete node features, namely {\method}~\citep{jin2021graph} and {\graphpca}~\citep{saerens2004principal}; (4) {\methodica}, (5) {\methodpca}, and (6) {\methodlda}, which incorporate {\method} with feature condensation by projecting the original features through a \textit{fixed feature condensation matrix} based on Independent Component Analysis (ICA)~\citep{comon1994independent}, Principal Component Analysis (PCA)~\citep{abdi2010principal}, and Linear Discriminant Analysis (LDA)~\citep{balakrishnama1998linear}, respectively.  
(7) Additionally, we consider a \textit{deep-learning} approach, namely the Variational Graph Autoencoder (VGAE)~\citep{kipf2017semi}, and utilizes its \textit{latent representation} as a condensed feature along with the GCond, denoted as {\methodvgae}.
To sum up, (1) does not involve any condensation process, while (2)-(3) implement node condensation through {\method} and GraphPCA, and (4)-(7) implement node condensation through {\method} and feature condensation by the corresponding algorithm. It is worth noting that for methods (4)-(7), where both feature size and node size condensation are involved, we perform feature condensation prior to node size condensation. 

\noindent\textbf{Evaluation.} 
To evaluate the effectiveness of condensed graphs, our approach involves several steps. Firstly, we obtain the condensed graph for the original training graph via each algorithm. Then we utilize the trained GNN model to infer labels for the test nodes on the entire graph. Note that the training graph corresponds to the complete graph in the transductive setting. The performance of the algorithm is then assessed by measuring the test accuracy.
For the {trainable condensed} graph, we retain $r_n \times N$ nodes for all algorithms, where $r_n$ represents the node condensation ratio, satisfying the condition $0 < r_n < 1$. Similarly, for baselines utilizing condensed features, the {learned condensed graph} has $r_n \times N$ nodes and $r_d \times D$ features. The parameter $r_d\ (0<r_d<1)$ signifies the ratio of {condensed} features to the original features. For the transductive setting, we utilize the full graph in the condensation process since the full graph is available in training. For the inductive setting, {\bf when only the training graph is available in training, we only condense the training graph. }

\noindent\textbf{Hyperparameter settings.}
{To implement} {\tinygraph}, we employ a 2-layer Simplified Graph Convolution (SGC)~\citep{wu2019simplifying} with $256$ hidden units as the GNN backbone. To capture the relationship between $\hat{\bf {A}}$ and $\hat{\bf {X}}$, we utilize a multi-layer perceptron (MLP) function, $g_{\bf{\Psi}}(\cdot)$. Specifically, for smaller graphs such as \cora and \citeseer, $g_{psi}(\cot)$ is a 3-layer MLP with $128$ hidden units in each hidden layer, while for larger graphs like \flickr and \reddit, we employ a 3-layer MLP with $256$ hidden units. We experiment with different {numbers of} training epochs from $\{600, 800, 1000\}$, and learning rates chosen from $\{1e-2, 5e-2, 1e-4, 5e-4, 1e-6, 1e-8\}$ for all the methods. Moreover, we set the value of $\gamma$ as $\{0.05, 0.05, 0.01, 0.01\}$ for \citeseer, \cora, \flickr, and \reddit, respectively. Additionally, we {set} $t_1$, $t_2$, and $t_{\boldsymbol{\theta}}$ as $\{20,15,10\}$ for \cora and \citeseer, and $\{20,10,20\}$ for \flickr and \reddit, respectively.

\subsection{Can {\tinygraph} achieve comparable performance with baselines using the original features?} 
\begin{table*}[!t]
\fontsize{9}{9}\selectfont
\setlength{\tabcolsep}{6pt}
\centering
\caption{The performance comparision of {\method}, {\graphpca}, {{{\methodica}}}, {{{\methodpca}}}, {\methodlda}, {\methodvgae}, and \colorbox{mygray}{\tinygraph}. \textbf{Acc} is the test accuracy on the original graph {without condensation}.
We report transductive performance on \cora, \citeseer, \arxiv, and inductive performance on \flickr, \reddit, with $r_d$ sets to $10\%$, $10\%$, $30\%$, $20\%$, and $30\%$, respectively. We marked the best result in bold.
}\label{table:baselines}
\vspace{-0.8em}
  \newcommand{\threecol}[1]{\multicolumn{3}{c}{#1}}
    \newcommand{\fivecol}[1]{\multicolumn{5}{c}{#1}}
  \newcommand{\fourcol}[1]{\multicolumn{4}{c}{#1}}
    \newcommand{\twocol}[1]{\multicolumn{2}{c}{#1}}
     \begin{tabular}{ccc|c|c|c|c|c|a}
    \toprule
        \texttt{Dataset}& Node Cond. &\twocol{Full Feature}&\fivecol{Condensed Feature} \\
       \cmidrule(l{4pt}r{4pt}){3-4} \cmidrule(l{4pt}r{4pt}){5-9}
                                                                           \textbf{Acc} (\%) & Ratio ($r_n$)&{\method}         &{\graphpca}        &{{{\methodica}}}        &{{{\methodpca}}}    &{{{\methodlda}}}    &{{{\methodvgae}}}    &{\tinygraph}           \\\midrule
\multirowcell{3}{ \cora\\\small{$81.3{\pm}0.4$}}        &$1.3\%$   & $80.9\pm3.2$      & $65.9\pm0.8$      &$72.8\pm4.2$            &$73.4\pm3.3$        &$75.4\pm2.8$        &$76.6\pm2.2$         &${\bf 79.3}\pm0.7$      \\
                                                        &$2.6\%$   & $80.6\pm2.7$      & $68.6\pm0.6$      &$74.3\pm1.4$            &$74.2\pm4.5$        &$76.2\pm2.4$        &$76.9\pm3.1$         &${\bf 80.1}\pm1.1$      \\    
                                                        &$5.2\%$   & $80.5\pm4.3$      & $70.2\pm1.2$      &$75.3\pm2.0$            &$75.4\pm3.5$        &$77.6\pm3.2$        &$77.8\pm4.1$         &${\bf 79.4}\pm1.5$      \\\midrule
\multirowcell{3}{\citeseer\\\small{$72.1{\pm}0.2$}}     &$0.9\%$   & $70.6\pm2.9$      & $60.1\pm1.3$      &$65.3\pm2.3$            &$65.8\pm1.1$        &$67.2\pm1.2$        &$67.9\pm2.8$         &${\bf 70.7}\pm0.6$      \\
                                                        &$1.8\%$   & $70.8\pm4.5$      & $63.4\pm0.8$      &$65.8\pm2.2$            &$65.3\pm2.8$        &$66.3\pm1.4$        &$64.9\pm3.5$         &${\bf 68.9}\pm1.7$      \\
                                                        &$3.6\%$   & $70.2\pm2.3$      & $58.6\pm1.8$      &$66.3\pm3.3$            &$66.8\pm2.2$        &$67.2\pm2.9$        &$65.8\pm2.6$         &${\bf 70.9}\pm0.8$      \\\midrule
\multirowcell{3}{\flickr\\\small{$47.3{\pm}0.1$}}       &$0.1\%$   & $47.1\pm4.1$      & $35.9\pm2.4$      &$41.7\pm1.9$            &$41.3\pm1.7$        &$42.5\pm2.1$        &$43.3\pm1.8$         &${\bf 45.5}\pm1.6$      \\
                                                        &$0.5\%$   & $47.2\pm1.5$      & $37.2\pm2.2$      &$43.6\pm1.8$            &$43.7\pm2.0$        &$43.9\pm1.1$        &$44.7\pm2.6$         &${\bf 46.7}\pm0.9$      \\
                                                        &$1.0\%$   & $47.2\pm3.4$      & $39.4\pm0.5$      &$42.8\pm3.7$            &$42.2\pm2.2$        &$44.5\pm2.8$        &$45.2\pm1.7$         &${\bf 46.9}\pm1.2$      \\\midrule
\multirowcell{3}{\reddit\\\small{$93.9{\pm}0.0$}}       &$0.05\%$  & $90.3\pm3.1$      & $84.2\pm1.9$      &$87.4\pm0.5$            &$87.3\pm1.4$        &$87.3\pm2.9$        &$87.3\pm1.6$         &${\bf 88.0}\pm0.7$      \\
                                                        &$0.1\%$   & $90.7\pm3.3$      & $85.1\pm4.3$      &${\bf 89.2}\pm3.7$      &$86.4\pm0.6$        &$87.4\pm1.7$        &$88.2\pm1.4$         &$89.1\pm1.1$      \\
                                                        &$0.2\%$   & $91.2\pm2.1$      & $86.8\pm2.1$      &$86.3\pm1.5$            &$86.8\pm2.5$        &$87.1\pm1.1$        &$88.2\pm4.7$         &${\bf 89.5}\pm1.3$       \\\midrule
\multirowcell{3}{\arxiv\\\small{$71.2{\pm}0.2$}}        &$0.05\%$  & $59.1\pm1.2$      & $54.9\pm2.2$      &$56.6\pm0.7$            &${\bf 58.9}\pm1.2$  &$57.6\pm1.1$        &$56.1\pm2.7$         &$58.8\pm0.9$      \\
                                                        &$0.25\%$  & $63.3\pm0.7$      & $52.2\pm1.5$      &$57.1\pm0.9$            &$60.0\pm0.9$        &$56.7\pm1.5$        &$59.2\pm3.2$         &${\bf 61.6}\pm1.3$      \\
                                                        &$0.5\%$   & $63.9\pm0.5$      & $55.4\pm2.1$      &$58.8\pm1.1$            &$61.3\pm1.2$        &$60.5\pm1.7$        &$60.5\pm3.6$         &${\bf 62.2}\pm0.8$       \\
                                                        \bottomrule
       \end{tabular}
  \end{table*}

\begin{table}[t]
\centering
\caption{The original test accuracy ($\%$) of various GNNs on original graphs. SAGE: GraphSAGE.}
\label{tab:original}\vspace{-0.8em}
\begin{tabular}{@{}l|c|c|c|c|cc@{}}
\toprule
            & SGC    & GCN     & SAGE      & APPNP   & GAT     \\ \midrule
\cora       & $81.4$ & $81.2$  & $81.2$    & ${\bf 83.1}$  & ${\bf 83.1}$  \\
\citeseer   & $71.3$ & $71.7$  & $70.1$    & ${\bf 71.8}$  & $70.8$  \\
\flickr     & $46.2$ & $47.1$  & $46.1$    & ${\bf 47.3}$  & $44.3$  \\
\reddit     & $93.5$ & $93.9$  & $93.0$    & ${\bf 94.3}$  & $91.0$  \\ 
\arxiv      & $71.4$ & ${\bf 71.7}$  & $71.5$    & $71.2$  & $71.5$  \\
\bottomrule
\end{tabular}
\vspace{-1em}
\end{table}
In this subsection, we present the experimental {results to validate} the performance of the condensed graph on node classification tasks, as shown in~\cref{table:baselines}. {\cref{tab:original} lists the performance of different GNN frameworks trained on the original graph.} {From} the results, it can be observed that
\noindent\Circled{\footnotesize 1}~\textbf{GNN trained with {\tinygraph} can achieve comparable performance to {\method} that trained with full features, at the same node condensation rate $r_n$.
}
Specifically, {\tinygraph} achieves test accuracies of up to $97.5\%$ on \citeseer and $98.7\%$ on \flickr, while reducing the graph size by $99.9\%$ and feature size up to $90\%$. This demonstrates the effectiveness of our proposed approach in condensing graph features while preserving important information. 
Overall, these results provide strong evidence for the utility of {\tinygraph} in tackling the challenges of training GNNs on large-scale graphs.
\subsection{Can {\tinygraph} archive better performance compared to structure-agnostic feature condensation baselines?} 
To answer this question, we compare {\tinygraph} with other baselines that use feature preprocesses as dimensionality reduction methods and present the results in~\cref{table:baselines}. From the results, we notice that
\noindent\Circled{\footnotesize 2}~\textbf{{\tinygraph} consistently obtains the best performance among all feature condensation baselines}.
This demonstrates the key role of structural information in feature condensation. At the same time, we notice that on \reddit, {\methodvgae} can achieve comparable performance with {\tinygraph}.
This {implies} that the original training graph structure of \reddit might not be useful. 
To verify this assumption, we train a GCN on the original \reddit dataset without using graph structure (i.e., setting ${\bf A}_\text{train}={\bf I}$), but {only use} the test graph structure for inference using the trained model. The obtained performance is $90.1\%$, which is indeed close to the original $93.9\%$, indicating that training without graph structure can still achieve comparable performance.
In addition, on the other three datasets, {\tinygraph} outperforms baselines that directly apply dimensionality reduction as a preprocessing method on the original feature, i.e., {{\methodica}}, {{\methodpca}}, {\methodlda}, and {\methodvgae}.
The reason lies {in the fact that these four} methods condense the feature independently from the graph structure. However, the correlation between feature and graph information is well known in literature~\citep{shalizi2011homophily,pfeiffer2014attributed}, overlooking the impact will lead to bad performance. {\tinygraph} jointly condenses features and nodes so it retains this relevance between features and structure.

\subsection{How many features are needed for {\tinygraph} to achieve equal performance with full-feature baselines?}
\begin{figure*}[t]
     \centering
     \subfloat[\cora]{{\includegraphics[height=0.145\linewidth]{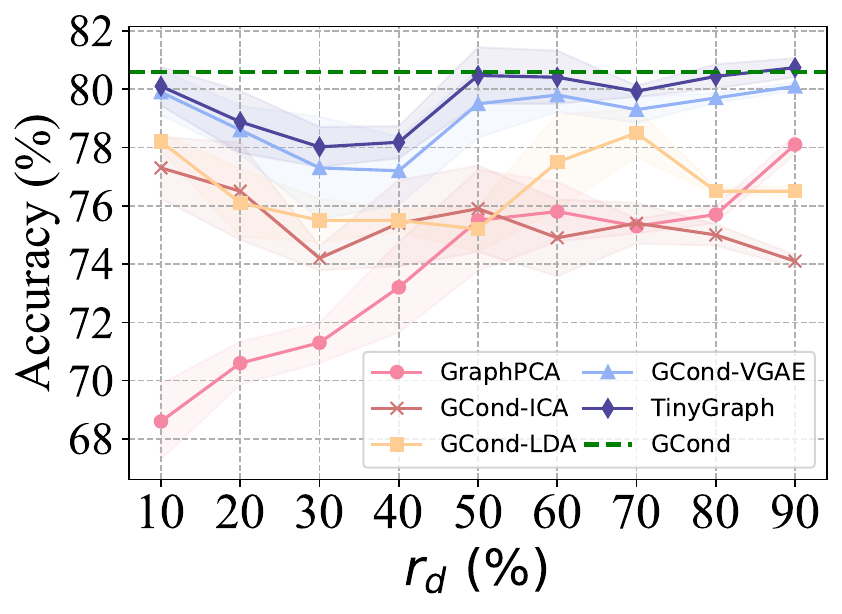} }}
     	\hspace{-4mm}
      \subfloat[\citeseer]{{\includegraphics[height=0.145\linewidth]{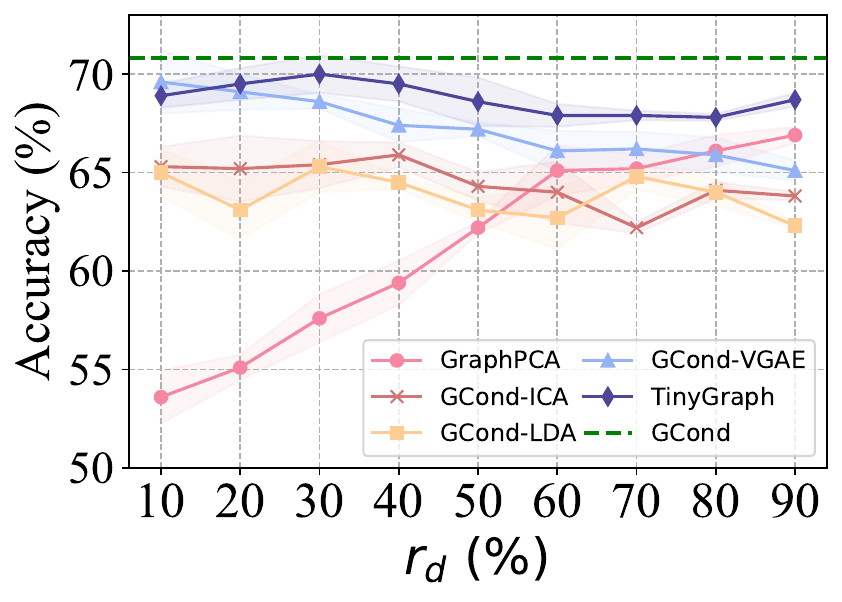} }}
      	\hspace{-4mm}
        \subfloat[\flickr]{{\includegraphics[height=0.145\linewidth]{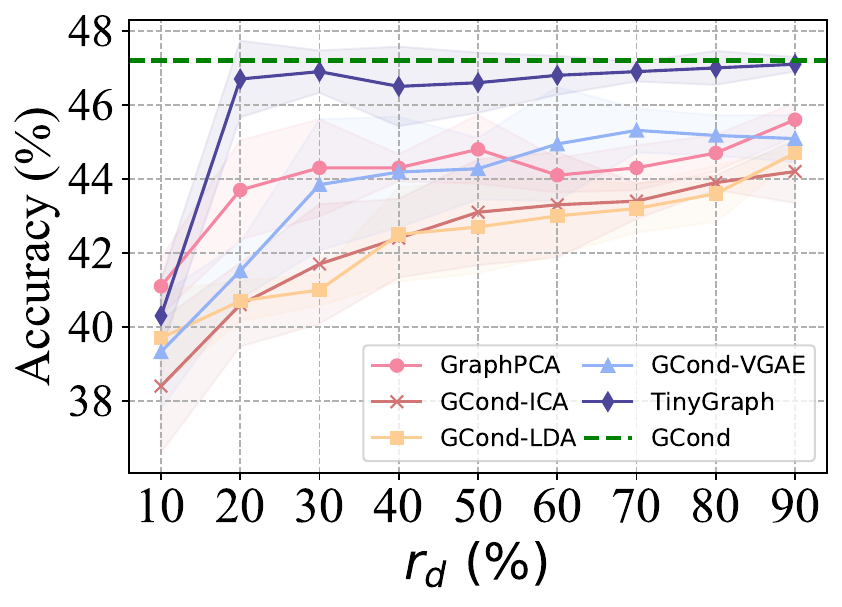} }}
        	\hspace{-4mm}
        \subfloat[\reddit]{{\includegraphics[height=0.145\linewidth]{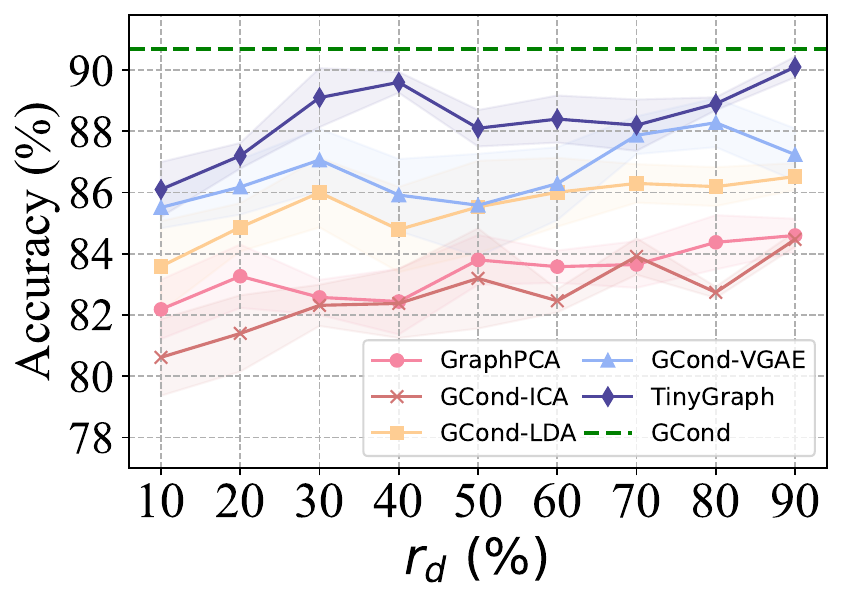} }}
        	\hspace{-4mm}
        \subfloat[\arxiv]{{\includegraphics[height=0.145\linewidth]{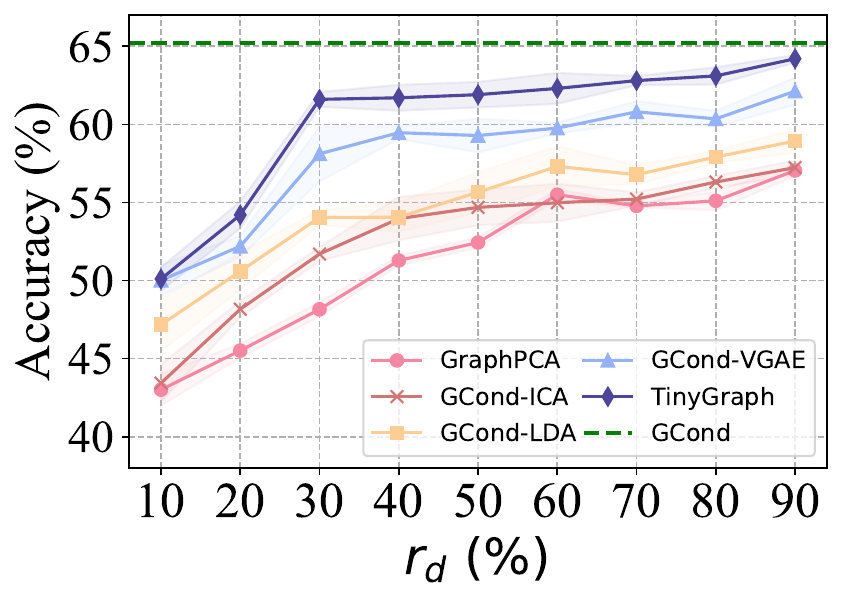} }}
        \vspace{-0.8em}
\caption{Performance comparison of our proposed method and baseline methods on various condensation ratios $r_d$, when $r_d$ is fixed to $2.6\%$, $1.8\%$, $0.5\%$, $0.1\%$, and $0.25\%$ on \cora, \citeseer, \flickr, \reddit, and \arxiv, respectively.}
\label{fig:feat_ratio}
\vskip -0.7em
\end{figure*}
To answer this question, we conducted experiments to demonstrate the effectiveness of {\tinygraph} for feature condensation across a range of feature condensing ratios $r_d$ from $10\%$ to $90\%$. We report our observations in \cref{fig:feat_ratio}, which shows that our proposed method outperforms the baseline methods across all feature condensing ratios tested. These results highlight the benefits of our proposed method for feature condensation, particularly in scenarios where the number of training nodes is limited, and feature dimensionality is high. We have the following two observations.
\noindent\Circled{\footnotesize 3}~\textbf{{\tinygraph} achieves equivalent performance while only utilizing $50\%$, $30\%$, $70\%$, and $90\%$ of the original feature size.} 
And even when $r_d$ is small enough, {\tinygraph} can still reach a high test accuracy compared to {\method} that uses full features. Specifically, $10\%$ in \cora, $10\%$ in \citeseer, $20\%$ in \flickr, $30\%$ in \reddit, and $30\%$ in \arxiv. the proposed {\tinygraph} is still able to perform the comparable performance ($95\%$ of the test accuracy) comparing to {\method} using the full-feature graph.
\noindent\Circled{\footnotesize 4}~\textbf{Larger feature size does not necessarily obtain better performance.}
The performance of {\tinygraph} varies with the number of training nodes in each dataset. When the number of training nodes is small, such as $35$ and $30$ in \cora and \citeseer, {\tinygraph} achieves the best performance with fewer features. However, for larger datasets like \flickr and \reddit, with $44,625$ and $153,932$ training nodes, respectively, {\tinygraph} performs best with a larger $r_d$. This behavior occurs because as the number of training nodes increases, a larger reduced dimensionality is needed to capture the complex relationships between the nodes. Therefore, for large datasets, it is necessary to use a larger $r_d$ to ensure optimal performance.

\subsection{What is the effect of GAT compared to other feature condensation {functions}? --- An Ablation study.}
\begin{table}[!t]
\setlength{\tabcolsep}{1.5pt}
\renewcommand{\arraystretch}{1.1}
\centering
\caption{Ablations on feature condensation function $f_{\bf \Phi}(\cdot)$.}
\vspace{-0.8em}
\label{tab:ablation}
\scalebox{0.90}
{
\begin{tabular}{@{}l|c|c|c|c|c@{}}
\toprule   
 Dataset               &\cora           &\citeseer      &\flickr        &\reddit        &\arxiv         \\ 
 ($r_n$)& ($2.6\%$)    & ($1.9\%$)      &($0.5\%$)      &  ($0.1\%$)    & ($0.25\%$)                     \\ \midrule
\tinygraphlin          & $78.1\pm3.2$   & $64.6\pm2.5$  & $41.2\pm3.1$  & $78.2\pm1.6$  & $43.3\pm1.0$   \\
\tinygraphmlp          & $78.9\pm2.6$   & $66.5\pm2.0$  & $42.3\pm2.3$  & $80.4\pm2.7$  & $55.1\pm2.2$   \\
\tinygraphgcn          & $79.2\pm1.9$   & $66.9\pm1.2$  & $46.3\pm1.5$  & $87.9\pm1.8$  & $57.8\pm0.9$   \\ 
\tinygraphone          & $79.8\pm1.5$   & $67.5\pm1.7$  & $45.2\pm1.8$  & $88.6\pm2.1$  & $59.0\pm1.7$   \\ \midrule
\tinygraph             & ${\bf 80.1}\pm1.1$   & ${\bf 68.9}\pm1.7$  & ${\bf 46.7}\pm0.9$  & ${\bf 89.1}\pm1.1$  & ${\bf 61.6}\pm1.3$   \\ 
\bottomrule
\end{tabular}
}
\vspace{-1.5em}
\end{table}
To address this question, we explored multiple implementations of {\tinygraph}, incorporating variations in the GAT function. These variants included: (1) a linear {feature condensation} function denoted as {\tinygraphlin}, (2) multi-layer perceptrons (MLPs) denoted as {\tinygraphmlp}, and (3) graph convolutional networks (GCNs) denoted as {\tinygraphgcn}. Additionally, we also consider another graph condensation framework based on one-step gradient matching~\citep{jin2022condensing}. Specifically, this method does not simulate the entire training trajectory of the original graph, which matches gradients at every epoch, but it only utilizes the gradient matching of the initial epoch. Based on this framework, we derive another variant of our method, (4) {\tinygraphone}.
\cref{tab:ablation} presents a comprehensive analysis, revealing that \noindent\Circled{\footnotesize 5}~\textbf{{\tinygraph} consistently outperforms all other variants across five datasets}, highlighting the efficacy of the chosen GAT function. Notably, on \cora and \citeseer, both characterized by a limited number of training nodes and abundant features, simple linear {feature condensation} functions and MLPs yield impressive performance.
The potential reason behind this observation is that the smaller-scale graphs (i.e., \cora and \citeseer) with large feature sizes are easy to condense, which has less requirement on the choice of {feature condensation} functions.
The rationale behind this observation lies in the relative ease of feature condensation due to the smaller scale of training nodes in \cora and \citeseer with ample features, i.e., ($D>>n$), rendering the {feature condensation} functions less stringent in their requirements.
Conversely, \flickr and \reddit, with their substantial number of training nodes and fewer features, pose a significantly more challenging scenario for the process of condensing structural awareness. 
Consequently, the utilization of structure-aware {feature condensation} functions, such as GCN and GAT, becomes imperative. 
\subsection{How does {\tinygraph} perform with various GNN models? --- A Generalizability Analysis. }
\begin{table}[!t]
\setlength{\tabcolsep}{3.5pt}
\centering
\caption{The performance of feature condensation baselines with different gradient matching (i.e. the test architecture). Avg. stands for the average test accuracy of MLP, GCN, GraphSAGE (SAGE), SGC, GAT, and APPNP. \tinygraph can work well on other gradient-matching architectures.}
\vspace{-0.8em}
\begin{tabular}{@{}c |l|cccccc|r}
\toprule   
                                                          & Methods          & MLP                           & GCN                           & SAGE                           & SGC                             & GAT                            & APPNP                          & Avg.                             \\ \midrule
\multirow{6}{*}{\rotatebox{90}{\fontsize{9}{9}\selectfont{\texttt{Cora}}}} 
                                                                 & \graphpca        &$63.0$                         &$66.9$                         &$66.1$                          &$64.4$                           &${\bf 68.6}$                          &$62.4$                          &$65.2$                           \\          
                                                                 & {\methodica}     &$67.5$                         &$72.6$                         &$70.6$                          &$69.8$                           &${\bf 74.3}$                          &$68.9$                          &$70.6$                           \\
                                                                 & {\methodpca}     &$68.5$                         &$72.4$                         &$69.7$                          &$69.4$                           &${\bf 74.2}$                          &$68.5$                          &$70.5$                           \\            
                                                                 & {\methodlda}     &$70.0$                         &$74.7$                         &$72.1$                          &$71.2$                           &${\bf 76.2}$                          &$70.3$                          &$72.4$                           \\               
                                                                 & {\methodvgae}    &$70.7$                         &$75.2$                         &$74.0$                          &$73.2$                           &${\bf 76.9}$                          &$71.0$                          &$73.5$                          \\          
                                                                 & {\cellcolor{mygray}\tinygraph}       & {\cellcolor{mygray}$78.9$}    & {\cellcolor{mygray}${\bf 79.8}$}    & {\cellcolor{mygray}$76.2$}     & {\cellcolor{mygray}$79.4$}      & {\cellcolor{mygray}${\bf 80.1}$}     & {\cellcolor{mygray}$79.3$}     & {\cellcolor{mygray}$79.0$}       \\\midrule 
\multirow{6}{*}{\rotatebox{90}{\fontsize{9}{9}\selectfont{\texttt{Citeseer}}}} 
                                                                & \graphpca         &$57.9$                         &${\bf 64.3}$                         &$60.1$                          &$62.7$                           &$63.4$                          &$59.1$                          &$60.9$                            \\         
                                                                & {\methodica}      &$59.5$                         &$64.7$                         &$62.8$                          &$65.2$                           &${\bf 65.8}$                          &$60.3$                          &$63.1$                             \\
                                                                & {\methodpca}      &$58.3$                         &$64.7$                         &$61.0$                          &$64.8$                           &${\bf 65.3}$                          &$59.4$                          &$62.2$                             \\
                                                                & {\methodlda}      &$59.6$                         &$65.8$                         &$61.9$                          &${\bf 66.7}$                           &$66.3$                          &$60.0$                          &$63.2$                            \\                      
                                                                & {\methodvgae}     &$57.9$                         &$64.3$                         &$60.5$                          &$64.4$                           &${\bf 64.9}$                          &$60.7$                          &$62.1$                            \\  
                                                                & {\cellcolor{mygray}\tinygraph}        & {\cellcolor{mygray}$67.5$}    & {\cellcolor{mygray}${\bf 68.9}$}    & {\cellcolor{mygray}$66.5$}     & {\cellcolor{mygray}$68.5$}      & {\cellcolor{mygray}${\bf 68.9}$}     & {\cellcolor{mygray}$68.0$}     & {\cellcolor{mygray}$68.2$}        \\\midrule       
\multirow{6}{*}{\rotatebox{90}{\fontsize{9}{9}\selectfont{\texttt{Flickr}}}} 
                                                                & \graphpca         &$33.0$                         &$36.3$                         &$35.5$                          &${\bf 37.9}$                           &$37.2$                          &$35.0$                          &$35.6$                            \\         
                                                                & {\methodica}      &$40.8$                         &$42.9$                         &$42.0$                          &${\bf 43.7}$                           &$43.6$                          &$40.9$                          &$42.2$                            \\
                                                                & {\methodpca}      &$39.8$                         &$42.6$                         &$42.1$                          &${\bf 44.2}$                           &$43.7$                          &$40.9$                          &$42.1$                            \\
                                                                & {\methodlda}      &$41.2$                         &$42.9$                         &$42.3$                          &$43.2$                           &${\bf 43.9}$                          &$41.4$                          &$42.5$                            \\                      
                                                                & {\methodvgae}     &$42.2$                         &$43.5$                         &$43.1$                          &$43.9$                           &${\bf 44.7}$                          &$42.0$                          &$43.2$                            \\  
                                                                & {\cellcolor{mygray}\tinygraph}        & {\cellcolor{mygray}$45.5$}    & {\cellcolor{mygray}$46.7$}    & {\cellcolor{mygray}$45.3$}     & {\cellcolor{mygray}$45.7$}      & {\cellcolor{mygray}$46.7$}     & {\cellcolor{mygray}${\bf 46.9}$}     & {\cellcolor{mygray}$46.1$}        \\\midrule    
\multirow{6}{*}{\rotatebox{90}{\fontsize{9}{9}\selectfont{\texttt{Reddit}}}} 
                                                                & \graphpca         &$81.9$                         &$84.0$                         &$83.5$                          &$84.6$                           &${\bf 85.1}$                          &$82.7$                          &$83.6$                            \\         
                                                                & {\methodica}      &$82.5$                         &$85.5$                         &$84.7$                          &$85.4$                           &${\bf 86.2}$                          &$83.3$                          &$84.6$                            \\
                                                                & {\methodpca}      &$83.7$                         &$85.0$                         &$84.9$                          &$85.7$                           &${\bf 86.4}$                          &$83.5$                          &$84.9$                            \\
                                                                & {\methodlda}      &$83.8$                         &$86.4$                         &$85.8$                          &${\bf 87.9}$                           &$87.4$                          &$84.4$                          &$85.8$                            \\                      
                                                                & {\methodvgae}     &$84.1$                         &$87.0$                         &${\bf 88.7}$                          &$87.5$                           &$88.2$                          &$85.0$                          &$86.4$                             \\  
                                                                & {\cellcolor{mygray}\tinygraph}        & {\cellcolor{mygray}$87.5$}    & {\cellcolor{mygray}${\bf 89.1}$}    & {\cellcolor{mygray}$88.7$}     & {\cellcolor{mygray}$87.9$}      & {\cellcolor{mygray}${\bf 89.1}$}     & {\cellcolor{mygray}$88.9$}     & {\cellcolor{mygray}$88.5$}         \\\midrule
\multirow{6}{*}{\rotatebox{90}{\fontsize{9}{9}\selectfont{\texttt{Arxiv}}}} 
                                                                & \graphpca         &$53.5$                         &$54.4$                         &$55.2$                          &$56.0$                           &$57.7$                          &$55.2$                          &${\bf 55.3}$                            \\  
                                                                & {\methodica}      &$55.1$                         &$56.9$                         &$57.3$                          &$57.0$                           &${\bf 57.8}$                          &${\bf 58.7}$                          &$57.1$                            \\
                                                                & {\methodpca}      &$54.8$                         &${\bf 58.9}$                         &$57.6$                          &$58.7$                           &$55.1$                          &$55.5$                          &$56.8$                            \\
                                                                & {\methodlda}      &$53.4$                         &$54.2$                         &$54.8$                          &$55.4$                           &$53.5$                          &${\bf 56.9}$                          &$54.7$                            \\                      
                                                                & {\methodvgae}     &$54.1$                         &${\bf 55.9}$                         &$55.7$                          &$54.4$                           &$55.3$                          &$54.6$                          &$55.0$                             \\  
                                                                & {\cellcolor{mygray}\tinygraph}        & {\cellcolor{mygray}$55.4$}    & {\cellcolor{mygray}$59.1$}    & {\cellcolor{mygray}$58.3$}     & {\cellcolor{mygray}${\bf 60.0}$}      & {\cellcolor{mygray}$59.7$}     & {\cellcolor{mygray}$58.2$}     & {\cellcolor{mygray}$58.5$}         \\\bottomrule                                                                

\end{tabular}
\label{tab:trans-models}
\end{table}
We demonstrate the generalizability of the \tinygraph in this experiment. Specifically, we evaluate the test performance by employing one GNN model for feature condensation while performing the gradient matching on other GNN architectures, including the default architecture SGC used in ~\cref{table:baselines}. The selected architectures for the {gradient matching graph neural network, i.e., $\text{GNN}_{\boldsymbol{\theta}}(\cdot)$}, include APPNP~\citep{klicpera2018predict-appnp}, GCN, SGC~\citep{wu2019simplifying}, GraphSAGE~\citep{hamilton2017inductive}, GAT~\citep{gat}, and MLP. The corresponding results are presented in \cref{tab:trans-models}. Our analysis of the table reveals that \Circled{\footnotesize 6}~\textbf{the condensed graph yields good performance even outside the scope it was optimized for}. This generalizability can be attributed to the similarity in filtering behaviors among these GNN models, as extensively investigated in prior studies~\citep{ma2021unified, zhu2021interpreting}.

\subsection{Will different feature condensation functions work with different GNN architectures for gradient matching?}
\label{subsec:archi}
\begin{figure}[t]
    \centering
    \subfloat[\cora]{\centering\includegraphics[width=0.245\textwidth]{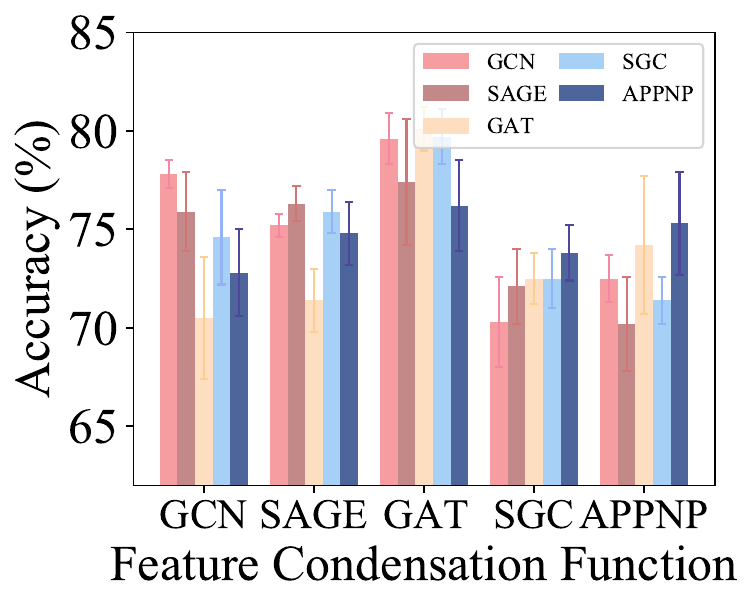}}
    \hspace{-2.8mm}
    \subfloat[\flickr]{\centering\includegraphics[width=0.245\textwidth]{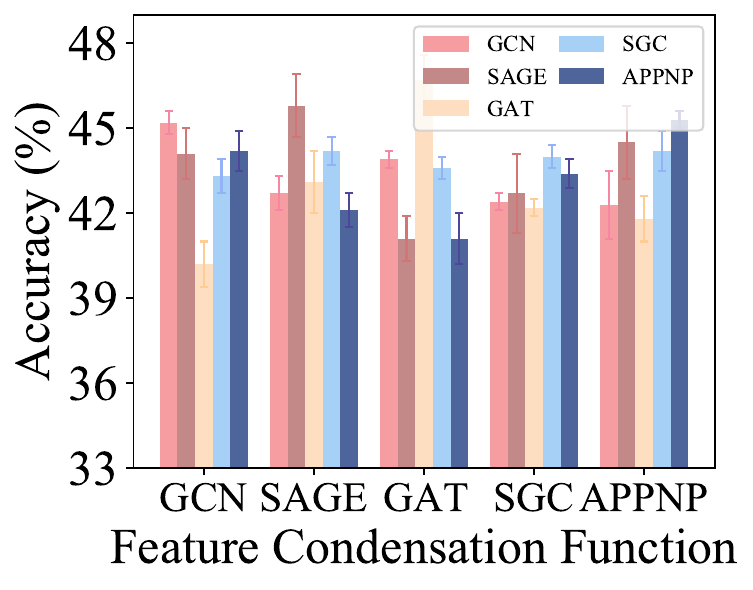}}
    \vspace{-0.8em}

    \caption{Cross-architecture performance is shown in test accuracy (\%). SAGE: GraphSAGE. Graphs condensed by different feature condensation functions all show strong transfer performance on other gradient-matching GNNs.}
    \label{fig:cross-valid}
\end{figure}
The condensed graph obtained from different feature condensation functions (which are instantiated as different graph neural networks) demonstrates its potential applicability to various GNN models. To investigate this applicability of {\tinygraph}, we evaluate the performance of a condensed graph produced by {one specific feature condensation function} for the {gradient matching utilizing alternative GNN architectures,} (i.e., {GCN, GraphSAGE, SGC, and APPNP}). The results are illustrated in \cref{fig:cross-valid}, where the x-axis denotes GNN architecture for {the feature} condensation, {and the y-axis represents the gradient matching performance measured by test accuracy of other GNN models.}
Our analysis reveals a noteworthy observation: \Circled{\footnotesize 7}~\textbf{the condensed graphs generated by different feature condensation functions exhibit promising applicability across different gradient matching GNNs}. 
This finding validates the versatility of the condensed graph and demonstrates the ability of {\tinygraph} to extract essential information from the original graph, resulting in a tiny condensed graph that retains practical utility for downstream tasks.

\subsection{
What are the specific statistics of the condensed graph V.S. the original graph? --- A Condensed Graph Analysis.}

\begin{table}[!t]
\centering
\fontsize{8}{9}\selectfont
\setlength{\tabcolsep}{2.0pt}
\caption{Statistics of condensed graphs and original graphs.
The percentages indicate the decrease of the statistic in the condensed graph compared to the one in the original graph. $\texttt{C-Seer}$ denotes the \citeseer dataset.}\label{tab:cond_stat}
\vspace{-0.8em}
\scalebox{0.960}
{
\begin{tabular}{l|l|ccccc|r} 
\toprule
                            &Methods                                            &\#Nodes                                            & \#Edges                                           & \#Features                                & Sparsity                                      & Storage                                   &Acc.                                               \\ \midrule
\multirow{3}{*}{\rotatebox{90}{\fontsize{7}{8}\selectfont{\texttt{Cora}}}}
                            & Full                                              & $2,708$                                             & $5,429$                                             &$1433$                                       &$99.9\%$                                      & $14.9$ MB                                    & $81.5\%$                                              \\
                            & {{\scriptsize{\textsf{TinyGraph}\xspace}}}        & $70$                                                & $2,128$                                             &$143$                                        &$14.4\%$                                      & 0.099 MB                                   & $80.1\%$                                              \\
                            & Decrease                                            &$97.4\%$	                                        &$60.8\%$	                                        &$90.0\%$           	                    &$85.6\%$                                      &$99.3\%$	                                &$1.5\%$                                            \\ \midrule
\multirow{3}{*}{\rotatebox{90}{\fontsize{7}{8}\selectfont{\texttt{C-Seer}}} }
                            & Full                                              & $3,327$                                             & $4,732$                                             &$3703$                                       &$99.9\%$                                      & $47.1$ MB                                    & $70.7\%$                                              \\
                            & {{\scriptsize{\textsf{TinyGraph}\xspace}}}        & $60$                                                & $1,454$                                             &$370$                                        &$20.55$                                       & $0.15$ MB                                    & $68.9\%$                                              \\
                            & Decrease                                            &$98.2\%$	                                        &$69.3\%$                                           &$90.0\%$   	       	                    &$79.4\%$                                      &$99.7\%$	                                &$2.5\%$                                            \\ \midrule
\multirow{3}{*}{\rotatebox{90}{\fontsize{7}{8}\selectfont{\texttt{Flickr}}} }
                            & Full                                              & $44,625$                                            & $218,140$                                           &$500$                                        &$99.9\%$                                      & $86.8$ MB                                    & $47.1\%$                                              \\
                            & {{\scriptsize{\textsf{TinyGraph}\xspace}}}        & $223$                                               & $3,788$                                             &$100$                                        &$84.8\%$                                      & $0.5$ MB                                     & $46.7\%$                                              \\
                            & Decrease                                            &$99.5\%$	                                        &$98.3\%$	                                        &$80.0\%$                                   &$15.1\%$                                      &$99.9\%$	                                &$0.8\%$                                            \\ \midrule
\multirow{3}{*}{\rotatebox{90}{\fontsize{7}{8}\selectfont{\texttt{Reddit}}}}
                            & Full                                              & $153,932$                                           & $10,753,238$                                        &$602$                                        &$99.9\%$                                      & $435.5$ MB                                   & $94.1\%$                                              \\
                            & {{\scriptsize{\textsf{TinyGraph}\xspace}}}        & $153$                                               & $301$                                               &$181$                                        &$97.4\%$                                      & $0.4$ MB                                     & $89.1\%$                                              \\
                            & Decrease                                            &$99.9\%$	                                        &$99.9\%$	                                        &$69.9\%$                                   &$2.47\%$                                      &$99.9\%$	                                &$5.3\%$                                          \\\midrule
\multirow{3}{*}{\rotatebox{90}{\fontsize{7}{8}\selectfont{\texttt{Arxiv}}}}
                            & Full                                              & $169,343$                                           & $1,166,243$                                        &$128$                                        &$99.9\%$                                      & $100.4$ MB                                   & $71.2\%$                                              \\
                            & {{\scriptsize{\textsf{TinyGraph}\xspace}}}        & $454$                                               & $3,354$                                               &$43$                                        &$97.8\%$                                      & $0.1$ MB                                     & $61.6\%$                                              \\
                            & Decrease                                            &$99.7\%$	                                        &$99.7\%$	                                        &$66.4\%$                                   &$2.10\%$                                      &$99.9\%$	                                &$13.5\%$                                          \\\bottomrule
\end{tabular}
}
\vspace{-1em}
\end{table}

In \cref{tab:cond_stat}, we present a comprehensive comparative analysis that examines several attributes distinguishing condensed graphs from their original counterparts. Our research findings reveal significant insights in the following areas.
Firstly, despite achieving comparable performance in downstream tasks, \Circled{\footnotesize 8}~\textbf{condensed graphs exhibit notable reductions in node count and feature dimensionality, thereby demanding significantly less storage capacity}. The feature size reduction is achieved while maintaining an accuracy level within acceptable tolerances.
Secondly, the condensed graphs demonstrate a lower degree of sparsity compared to their larger counterparts. This observation arises due to the inherent challenge of maintaining the original level of sparsity in graphs that are significantly smaller in scale. Preserving the original sparsity in condensed graphs would result in minimal inter-node connections, which may hinder the effectiveness of the condensed graph.

\subsection{How is the quality of the learned feature by {\tinygraph}? --- A Visualization of Condensed Features.}
\begin{figure}[]
\begin{center}
\setlength{\abovecaptionskip}{-0.1cm}
\centering
\includegraphics[width=0.95\linewidth]{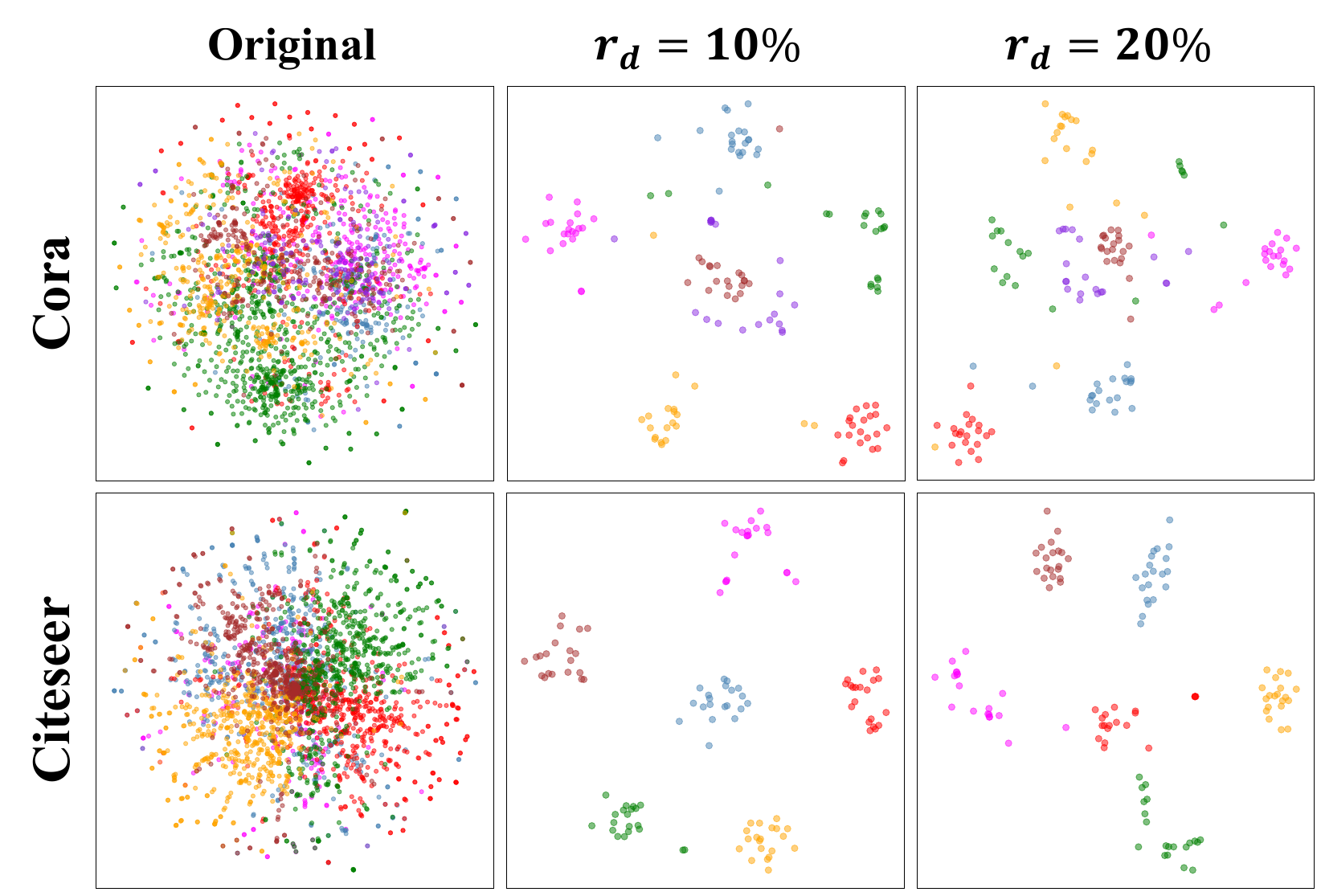}
\end{center}
\vspace{-0.8em}
\caption{t-SNE visualization of original node features and condensed node features learned by {\tinygraph}, on \cora and \citeseer, with $r_d=10\%$ and $r_d=20\%$ for both datasets. Different colors represent different classes. The distinctly separable clusters in the t-SNE of condensed features demonstrate the discriminative capability of {\tinygraph}.}
\label{fig:tsne}
\vspace{-1.2em}
\end{figure}
To address this question, we conducted a visual analysis of the t-SNE~\citep{van2008visualizing} embeddings derived from both the original features and the condensed features of the \cora and \citeseer datasets in \cref{fig:tsne}. The condensation ratios employed were set at $10\%$ and $20\%$, denoted as $r_d=10\%$ and $r_d=20\%$, respectively. \cref{fig:tsne} shows that the condensed features learned by our proposed {\tinygraph}, exhibit distinct separable clusters while the original node feature is mixed together. This finding provides compelling evidence of {\tinygraph}'s capability to learn highly discriminative features. 
Remarkably, upon closer examination of both datasets, the t-SNE plots reveal that the embeddings obtained with a condensation ratio of $10\%$ demonstrate more discernible patterns compared to those obtained with a condensation ratio of $20\%$. This observation suggests that a smaller condensation ratio can yield even more discriminative features. The underlying reason lies in {\tinygraph}'s ability to map nodes from different classes into distinct communities and thereby maximize the dissimilarity between the nodes within the same class.

\section{Conclusion}\label{sec:conclusion}

In this work, we introduce a new joint graph condensation framework, named {\tinygraph}. Unlike traditional approaches that only focus on condensing nodes, {\tinygraph} is designed to condense both nodes and features of a large-scale graph. Despite the condensation, it meticulously preserves the original graph information. To optimize {\emph{the trainable condensed graph}}, we employ a gradient matching strategy, while a structure-aware dimensionality function is used to maintain the integrity of the graph structure. This dual condensation allows {\tinygraph} to achieve high test accuracies across various datasets, demonstrating its efficiency. {\tinygraph} has potential applications in training Graph Neural Networks on large-scale graphs. This is particularly relevant for graphs with massive nodes and high-dimensional features, where computational resources, such as memory and time, are limited.

\appendix
\subsection{Implementation Details}
\subsubsection{Hyperparameter Setting}
We present our hyperparameter configuration, encompassing three key stages: node condensation, feature condensation, and evaluation. Additionally, the hyperparameters are detailed in both \cref{tab:trans-models} and \cref{fig:cross-valid}.

\noindent\textbf{Node Condensation:} 
We compare our \tinygraph on two main methods, \graphpca, and \method.

In the context of \tinygraph, our approach involves the application of a $2$-layer Simplified Graph Convolutional Network (SGC) with $256$ hidden units, serving as the Graph Neural Network (GNN) for gradient matching. The function $g_{\Phi}$, representing the relationship between ${\bf A'}$ and ${\bf X'}$, is implemented as a multi-layer perceptron (MLP). Specifically, we employ a 3-layer MLP with $128$ hidden units for small graphs (\cora and \citeseer) and $256$ hidden units for larger graphs (\flickr, \reddit, and \arxiv). We also explore various training epochs for \tinygraph, ranging from $\{400, 600, 1000\}$. For \graphpca, we leverage the implementation available at~\footnote{https://github.com/brandones/graphpca}, configuring `add\_supernode' as False and `eigendecomp\_strategy' as `exact'. In the case of~\method, we adopt a parameter setting similar to \tinygraph, as previously elucidated.
\begin{table}[b]
\vspace{-1em}
\centering
\caption{Running time of~\tinygraph for $20$ epochs (in seconds), with $r_d$ set to $10\%, 30\%, 30\%$ on \cora, \flickr, and \arxiv.}
\label{tab:time}
\begin{tabular}{@{}l|c|c|c|c|c@{}}
\toprule
$r_n (\%)$          & 0.2 & 0.5  & 1 &5 &10 \\\midrule
\cora        & 5.7& 11.0& 21.8  & 24.6  & 36.3  \\
\flickr & 87.4  & 112.2  & 167.9 &214.1 &302.5  \\
\arxiv & 201.7     & 231.8 & 309.4  &467.5 &642.1
 \\ \bottomrule
\end{tabular}
       \vspace{-1em}
\end{table}
\begin{table*}[t]
\fontsize{9}{9}\selectfont
\setlength{\tabcolsep}{7pt}
\renewcommand{\arraystretch}{1.2}
\centering
\caption{Performance comparison with {\tinygraphx} (solely learning the feature matrix), {\methodpcaf} and {\methodicaf} (with a reversed order of condensation processes for {\methodpca} and {\methodica}—specifically, first condensing nodes and then condensing features using PCA and ICA). Additionally, we evaluated the performance of {{{\tinygraphthree}}} (with the layer set to $3$), {{{\tinygraphote}}} (with hidden units set to $128$), and {\tinygraphhfot} (with hidden units set to $512$). We report the test accuracy on the original graph without any condensation.
For transductive performance, we assessed the models on \cora, \citeseer, \arxiv, and for inductive performance, we used \flickr, \reddit. The condensation ratio $r_d$ was set to $10\%$ for \cora and \citeseer, $30\%$ for \arxiv, $20\%$ for \flickr, and $30\%$ for \reddit, respectively.
}\label{table:additional}
  \newcommand{\threecol}[1]{\multicolumn{3}{c}{#1}}
    \newcommand{\fivecol}[1]{\multicolumn{5}{c}{#1}}
  \newcommand{\fourcol}[1]{\multicolumn{4}{c}{#1}}
    \newcommand{\twocol}[1]{\multicolumn{2}{c}{#1}}
     \begin{tabular}{c|c|c|c|c|c|c|c}
    \toprule
                                                                              &$r_n (\%)$   &{\tinygraphx}    &{\methodpcaf}     &{{{\methodicaf}}}   &{{{\tinygraphthree}}}  &{{{\tinygraphote}}}     &{{{\tinygraphhfot}}}   \\\midrule
\multirowcell{3}{\rotatebox{90}{\fontsize{9}{9}\selectfont{\texttt{Cora}}}}   &$1.3$        &${ 76.6}\pm1.1$  &${ 75.1}\pm0.7$   &${ 75.3}\pm1.2$     &${ 79.1}\pm1.2$        &${\bf 79.5}\pm0.8$         &${ 78.5}\pm1.3$        \\
                                                                              &$2.6$        &${ 78.3}\pm0.9$  &${ 75.4}\pm1.1$   &${ 76.8}\pm0.7$     &${\bf 79.4}\pm1.3$        &${\bf 79.4}\pm1.2$         &${ 79.3}\pm1.0$        \\    
                                                                              &$5.2$        &${ 77.5}\pm1.6$  &${ 76.2}\pm1.2$   &${ 76.4}\pm1.0$     &${\bf 79.2}\pm1.1$        &${\bf 79.2}\pm2.3$         &${ 78.3}\pm1.1$        \\\midrule
\multirowcell{3}{\rotatebox{90}{\fontsize{9}{9}\selectfont{\texttt{C-Seer}}}} &$0.9$        &${\bf 71.5}\pm1.3$  &${ 66.2}\pm1.4$   &${ 66.7}\pm0.9$     &${ 69.6}\pm1.2$        &${ 69.8}\pm1.1$         &${ 69.1}\pm0.8$        \\
                                                                              &$1.8$        &${\bf 69.7}\pm1.4$  &${ 66.1}\pm1.2$   &${ 67.5}\pm1.4$     &${ 69.1}\pm0.9$        &${ 69.2}\pm1.4$         &${ 68.4}\pm1.1$        \\
                                                                              &$3.6$        &${\bf 71.7}\pm1.2$  &${ 67.4}\pm1.7$   &${ 67.6}\pm1.3$     &${ 69.2}\pm1.4$        &${ 69.0}\pm0.9$         &${ 69.2}\pm1.7$        \\\midrule
\multirowcell{3}{\rotatebox{90}{\fontsize{9}{9}\selectfont{\texttt{Flickr}}}} &$0.1$        &${ 43.6}\pm1.5$  &${ 41.2}\pm1.5$   &${ 41.1}\pm0.8$     &${ 43.2}\pm0.7$        &${ 46.1}\pm1.2$         &${\bf 46.2}\pm1.7$        \\
                                                                              &$0.5$        &${ 44.8}\pm1.1$  &${ 42.6}\pm1.6$   &${ 42.2}\pm1.1$     &${ 44.6}\pm1.1$        &${\bf 46.4}\pm1.4$         &${ 45.3}\pm1.2$        \\
                                                                              &$1.0$        &${ 44.5}\pm1.3$  &${ 43.1}\pm1.2$   &${ 44.6}\pm1.2$     &${ 45.2}\pm0.9$        &${\bf 46.9}\pm1.2$         &${ 46.4}\pm1.5$        \\\midrule
\multirowcell{3}{\rotatebox{90}{\fontsize{9}{9}\selectfont{\texttt{Reddit}}}} &$0.05$       &${\bf 89.7}\pm2.0$  &${ 84.3}\pm0.8$   &${ 84.8}\pm1.5$     &${ 87.4}\pm2.1$        &${ 89.1}\pm0.9$         &${ 89.3}\pm2.2$        \\
                                                                              &$0.1$        &${ 90.2}\pm2.2$  &${ 85.2}\pm1.1$   &${ 85.7}\pm0.9$     &${\bf 90.4}\pm1.4$        &${ 88.4}\pm2.2$         &${ 89.2}\pm2.1$        \\
                                                                              &$0.2$        &${\bf 90.8}\pm1.1$  &${ 86.0}\pm1.2$   &${ 86.2}\pm1.4$     &${ 89.5}\pm1.3$        &${ 89.0}\pm1.9$         &${ 90.1}\pm2.3$        \\\midrule
\multirowcell{3}{\rotatebox{90}{\fontsize{9}{9}\selectfont{\texttt{Arxiv}}}}  &$0.05$       &${\bf 58.3}\pm0.6$  &${ 55.8}\pm1.2$   &${ 56.0}\pm0.9$     &${ 57.4}\pm1.1$     &${ 57.6}\pm1.2$         &${ 56.2}\pm2.0$         \\
                                                                              &$0.25$       &${ 58.3}\pm1.3$  &${ 55.1}\pm1.4$   &${ 58.1}\pm1.9$     &${ 59.2}\pm0.7$        &${ 58.7}\pm1.4$         &${\bf 59.4}\pm1.1$        \\
                                                                              &$0.5$        &${ 60.1}\pm1.8$  &${ 56.3}\pm1.5$   &${ 57.9}\pm1.3$     &${ 60.5}\pm0.9$        &${\bf 60.7}\pm1.6$      &${ 60.3}\pm1.8$        \\\bottomrule
       \end{tabular}
       \vspace{-1em}
  \end{table*}
We performed hyperparameter tuning for all methods by adjusting the learning rate, considering values within the range of $\{0.1, 0.01, 0.001, 0.0001\}$. Additionally, we assigned specific values to the parameter $\delta$ for different datasets: $0.05$ for \citeseer, $0.05$ for \cora, $0.01$ for \arxiv, $0.01$ for \flickr, and $0.01$ for \reddit.

Addressing the condensation ratio choices, our discussion is divided into two sections. The first section focuses on transductive datasets, posing challenges due to their low labeling rates. In the case of \cora and \citeseer, with labeling rates of only 5.2\% and 3.6\%, respectively, we expressed condensation ratios as percentages of the labeling rates. For \cora, we selected $r$ values of $\{25\%, 50\%, 100\%\}$, resulting in condensation ratios of $\{1.3\%, 2.6\%, 5.2\%\}$. Similarly, for \citeseer, $r$ values of $\{0.9\%, 1.8\%, 3.6\%\}$ yielded the desired condensation ratios. For \arxiv, with a labeling rate of $53\%$, we set $r$ to $\{0.1\%, 0.5\%, 1\%\}$ of this rate, resulting in condensation ratios of $\{0.05\%, 0.25\%, 0.5\%\}$. The second section contains inductive datasets, where all nodes in the training graphs are labeled. Here, we selected different $r$ values to ensure the desired condensation ratios. Specifically, for \flickr, we chose $\{0.1\%, 0.5\%, 1\%\}$, and for \reddit, $\{0.05\%, 0.1\%, 0.2\%\}$ were selected as our $r$ values.

\noindent\textbf{Feature Condensation:}
In this analysis, we primarily assess the efficacy of the suggested joint condensation technique by juxtaposing it against conventional dimensionality reduction methods such as ICA, PCA, and LDA, along with the deep learning approach VGAE. Note that, unless otherwise specified, default parameters are employed for all functions. Specifically, for ICA, PCA, and LDA, the `PCA' and `FastICA' functions within the `sklearn.decomposition' package\footnote{\href{https://scikit-learn.org/stable/modules/generated/sklearn.decomposition.FastICA.html}{sklearn.decomposition.FastICA}} are utilized. For VGAE, we use a public Pytorch implementation\footnote{\href{https://github.com/DaehanKim/vgae_pytorch}{https://github.com/DaehanKim/vgae\_pytorch}}. In our approach, we adopt the weight initialization method outlined in previous studies~\citep{glorot2010understanding}. The training process consists of $200$ iterations, employing the Adam optimizer~\citep{kingma2014adam} with a learning rate set to $0.01$. Across all experiments, a hidden layer of dimension $32$ and latent variables of dimension $16$ are utilized.

\noindent{\textbf{Testing:}
During the evaluation process, we adhere to specific parameter configurations for training various GNNs. We fix the dropout rate at $0$ and set the weight decay to $0.0005$. The GAT model undergoes 3000 training epochs, while other models are trained for $600$ epochs. Throughout, the initial learning rate is consistently maintained at $0.01$.

\noindent{\textbf{Settings for~\cref{tab:trans-models} and~\cref{fig:cross-valid}:}
For both the condensation and evaluation stages, we maintain a constant GNN depth of $2$. During the condensation stage, the weight decay is set to $0$, dropout to 0, and training epochs to $1000$. Subsequently, in the evaluation stage, we adjust the weight decay to $0.0005$, keep dropout at $0$, and set the training epochs to 600.
\subsection{Efficiency}
We also evaluate the efficiency of the proposed method. Specifically, we first analyze the time complexity of {\tinygraph} and then compare the run time of our method with baselines.

\subsubsection{Run Time Analysis}
We present the runtime analysis of our proposed method across various condensation rates. Specifically, we explore condensation rates within the range of $\{0.1\%, 0.5\%, 1\%\}$ for \arxiv and $\{1\%, 5\%, 10\%\}$ for \cora. The execution times for $30$ epochs on a NVIDIA RTX A4000 GPU are detailed in~\cref{tab:time}. 

\subsection{Additional Experiments}
We conducted additional experiments focusing on three key aspects. First, we investigated the sole learning of the feature matrix. Second, we explored the effects of reversing the order of node and feature condensation processes. Third, we delved into neural architecture search. It's worth noting that for the third aspect, we previously examined various GNN architectures in~\cref{subsec:archi}. In this context, our current discussion primarily centers on the parameter modifications applied to the default architecture, namely the two-layer SGC with $128$ hidden units. We contemplated altering it by either introducing three layers or adjusting the hidden units to $128$ or $512$, referred to as {\tinygraphthree}, {\tinygraphhfot}, or {\tinygraphote} respectively.

From~\cref{table:additional}, we have the following observations:
\begin{itemize}
\item Learning $\hat{\bf {X}}$ and $\hat{\bf {A}}$ concurrently allows direct absorption of graph structure into learned features, reducing the need to consistently distill graph properties. This approach maintains good generalization performance from features.
\item Using a three-layer SGC or employing $512$ hidden units, such as {\tinygraphthree} and {\tinygraphhfot}, results in inferior performance compared to the default two-layer $256$ SGC. This aligns with the intuitive understanding that deeper and wider GNNs can lead to over-smoothing. Regarding {\tinygraphote}, reducing the hidden units to $128$ yields slightly worse but closely comparable performance to the default {\tinygraph}, indicating that decreasing hidden units has minimal impact on performance.
\item Reversing the sequence of node and feature condensation, where the initial condensation of the original graph followed by condensing features on the reduced graph results in poor performance. The issue arises from the asynchronous learning of node and feature condensation, causing a lack of synchronized information in the subsequent feature condensation step.
\end{itemize}

\bibliographystyle{IEEEtran}
\bibliography{ref.bib}

\end{document}